\documentclass{article}

\usepackage{microtype}
\usepackage{graphicx}
\usepackage{subfigure}
\usepackage{booktabs} 
\usepackage{multirow}
\usepackage{xcolor}
\usepackage{colortbl}
\usepackage{enumitem}
\usepackage{tcolorbox}
\tcbuselibrary{listings,skins,breakable}

\usepackage{hyperref}

\usepackage[preprint]{icml2026}

\usepackage{amsmath}
\usepackage{amssymb}
\usepackage{mathtools}
\usepackage{amsthm}
\usepackage{algorithm}
\usepackage{algorithmic}

\usepackage[capitalize,noabbrev]{cleveref}

\theoremstyle{plain}

\theoremstyle{definition}

\theoremstyle{remark}

\newcommand{\dsName}{\textsc{GPSBench}}

\newtcolorbox{llmprompt}[1][]{
  enhanced,
  breakable,
  width=\columnwidth,
  colback=gray!5,
  colframe=gray!40,
  boxrule=0.5pt,
  left=4pt,
  right=4pt,
  top=4pt,
  bottom=4pt,
  fontupper=\ttfamily\footnotesize,
  #1
}

\newtcolorbox{systemprompt}[1][]{
  enhanced,
  breakable,
  width=\columnwidth,
  colback=blue!3,
  colframe=blue!30,
  boxrule=0.5pt,
  left=4pt,
  right=4pt,
  top=4pt,
  bottom=4pt,
  fontupper=\ttfamily\footnotesize,
  title={\footnotesize\sffamily System Prompt},
  #1
}

\newtcolorbox{taskprompt}[1][]{
  enhanced,
  breakable,
  width=\columnwidth,
  colback=green!3,
  colframe=green!30,
  boxrule=0.5pt,
  left=4pt,
  right=4pt,
  top=4pt,
  bottom=4pt,
  fontupper=\ttfamily\footnotesize,
  title={\footnotesize\sffamily Task Prompt},
  #1
}

\begin{document}

\twocolumn[
  \icmltitle{GPSBench: Do Large Language Models Understand GPS Coordinates?}



  \icmlsetsymbol{equal}{*}

  \begin{icmlauthorlist}
    \icmlauthor{Thinh Hung Truong}{yyy}
    \icmlauthor{Jey Han Lau}{yyy}
    \icmlauthor{Jianzhong Qi}{yyy}
  \end{icmlauthorlist}

  \icmlaffiliation{yyy}{University of Melbourne}

  \icmlcorrespondingauthor{Thinh Hung Truong}{truonghungthinh21@gmail.com}

  \icmlkeywords{Geospatial, LLM, GPS}

  \vskip 0.3in
]



\printAffiliationsAndNotice{}  

\begin{abstract}
Large Language Models (LLMs) are increasingly deployed in applications that interact with the physical world, such as navigation, robotics, or mapping, making robust geospatial reasoning a critical capability. Despite that, LLMs’ ability to reason about GPS coordinates and real-world geography remains underexplored. We introduce \dsName, a dataset of 57,800 samples across 17 tasks for evaluating geospatial reasoning in LLMs, spanning geometric coordinate operations (e.g., distance and bearing computation) and reasoning that integrates coordinates with world knowledge. Focusing on intrinsic model capabilities rather than tool use, we evaluate 14 state-of-the-art LLMs and find that GPS reasoning remains challenging, with substantial variation across tasks: models are generally more reliable at real-world geographic reasoning than at geometric computations.
Geographic knowledge degrades hierarchically, with strong country-level performance but weak city-level localization, while robustness to coordinate noise suggests genuine coordinate understanding rather than memorization. We further show that GPS-coordinate augmentation can improve in downstream geospatial tasks, and that finetuning induces trade-offs between gains in geometric computation and degradation in world knowledge. Our  dataset and reproducible code are available at \url{https://github.com/joey234/gpsbench/}
\end{abstract}


\section{Introduction}
\label{sec:intro}

Large Language Models (LLMs) have achieved strong performance across diverse reasoning tasks, from mathematical problem-solving \citep{cobbe2021gsm8k,hendrycks2021measuring} to coding \citep{chen2021evaluatinglargelanguagemodels}.
As LLMs are increasingly integrated into location-aware applications such as navigation assistants, geographic information systems, travel planning, and emergency response, understanding their capabilities in Global Positioning System (GPS) coordinates and spatial reasoning becomes essential. Consider the range of geographic queries users might pose: ``How far is it from Tokyo to Sydney?'', ``What city is at coordinates 48.86°N and  2.35°E?'', or ``Is Portland north or south of Toronto?''. These questions require not just factual recall, but genuine spatial reasoning over the Earth's coordinate system such as computing great-circle distances, understanding coordinate-to-place mappings, and reasoning about relative positions across the globe.

Despite growing interest in LLM spatial reasoning \citep{liu2022vsr, chen2024spatialvlm, yamada2024evaluating}, existing benchmarks focus primarily on small-scale perceptual tasks (e.g., ``is the cup to the left of the book?'') or visual spatial relations in images and 3D scenes \citep{kamath-etal-2023-whats,marcu2023lingoqa}. 
These evaluations largely focus on small-scale, tabletop, or highly-controlled environments (e.g., rooms, grids, or toy maps), where spatial extent is limited and uncertainty is minimal. As a result, they provide limited insight into LLMs’ ability to acquire and manipulate survey knowledge in realistic geographic settings, where spatial reasoning must operate over continuous space, noisy observations, spherical geometry, and complex real-world constraints. This hypothesis is further supported by \citet{yamada2024evaluating} and \citet{yang2024think}, revealing that models achieve reasonable local spatial awareness but struggle with global spatial representations.

On the other hand, geographic spatial reasoning involves dealing with global-scale GPS coordinates, distances spanning thousands of kilometers, and requiring integration of world knowledge about countries, cities, and terrains, which remains underexplored. This gap is significant: geographic reasoning differs fundamentally from tabletop spatial reasoning in scale (meters vs.\ thousands of kilometers), geometry (planar vs.\ spherical), and knowledge requirements (perceptual vs.\ world knowledge). A model that correctly reasons about object positions on a table may struggle when it is asked to compute distances or identify the location of a given GPS coordinate.

To address this gap, we propose  \textbf{\dsName}, the first comprehensive benchmark for evaluating GPS and location-based reasoning in LLMs. \dsName\ comprises of 57,800 samples across 17 tasks organized into two tracks: geometric coordinate operations (such as distance, bearing, coordinate transformations, and spherical geometry) requiring mathematical reasoning but no world knowledge, and applied geographic reasoning (such as coordinate-to-place mapping, spatial relationships, and pattern recognition) requiring integration with real-world knowledge. Tasks are grounded in the landmark-route-survey framework from spatial cognition \citep{siegel1975development}. All samples are derived from the GeoNames database~\cite{GeoNames2026} covering 18,196 locations across six continents, with ground truth computed via geodetic formulae or derived from authoritative databases. Unlike tool-augmented benchmarks \citep{krechetova2025geobenchx, zhang2025geoanalystbench}, \dsName\ measures intrinsic capabilities: what LLMs know and can compute from parameters alone, and which matters for latency-sensitive, offline, or privacy-restricted deployments. It also reveals training data biases that may be overlooked by tool use.

Our contributions are:

\begin{itemize}
    \item \textbf{\dsName}: A large-scale benchmark of 57,800 samples across 17 tasks organized into Pure GPS and Applied tracks, with tasks grounded in the landmark-route-survey framework from spatial cognition \citep{siegel1975development}.

    \item A systematic evaluation of 14 state-of-the-art LLMs, analyzing performance across task types, model families and sizes, and geographic granularities. We show that GPS reasoning varies strongly across tasks: models perform reasonably on basic geometric
operations but struggle with complex spherical geometry
and fine-grained place association.
    Geographic knowledge degrades hierarchically, with strong country-level identification but weak city-level localization, while robustness to coordinate noise suggests genuine coordinate understanding rather than memorization. We further demonstrate that augmenting downstream benchmarks with GPS coordinates yields substantial improvements, and that finetuning introduces trade-offs, improving geometric computation at the expense of real-world geographic knowledge.
\end{itemize}


\section{Related work}
\label{sec:related}

\subsection{Spatial Cognition Framework}

The psychology of spatial knowledge acquisition provides a foundational \emph{landmark-route-survey} taxonomy for understanding geographic reasoning \citep{siegel1975development}. This framework distinguishes three hierarchical but complementary forms of spatial knowledge. \emph{Landmark knowledge} involves the recognition and recall of salient environmental features (e.g., buildings, intersections, natural features) without explicit encoding of metric distances or spatial relations. \emph{Route knowledge} captures ordered sequences of actions or paths that connect landmarks, typically represented as procedural and egocentric knowledge tied to a specific traversal experience (e.g., turn left at the church, then walk straight to the bridge). \emph{Survey knowledge} represents an allocentric, map-like understanding of space, encoding global layout, metric distances, and directional relationships, thereby enabling flexible reasoning such as shortcut discovery, detour planning, and novel route inference.
These knowledge types contribute to geospatial reasoning capabilities and are acquired through different learning modalities and exhibit distinct strengths and limitations. For example, \citet{thorndyke1982differences} demonstrate that navigation-based learning preferentially supports route knowledge, whereas map-based learning facilitates survey knowledge. 
We adopt this framework as a foundation for building a comprehensive geospatial understanding benchmark.

\subsection{Geospatial Benchmarks}

Evaluating LLMs on geospatial tasks has emerged as an active research area, with benchmarks spanning geographic knowledge assessment, spatial reasoning, coordinate-based computation, and tool-augmented workflows.

\paragraph{Geographic knowledge and bias}
Several benchmarks evaluate LLM factual geographic knowledge. WorldBench~\citep{moayeri2024worldbench} tests country-level indicator (i.e., statistics such as population or carbon dioxide emissions) recall, revealing 1.5$\times$ higher error rates for Sub-Saharan Africa versus North America. \citet{manvi2024geographically} demonstrate systematic biases against low-socioeconomic regions. GeoLLM \citep{manvi2024geollm} shows that LLMs embed geospatial knowledge which is extractable through prompting with OpenStreetMap data. GeoGLUE \citep{li2023geoglue} evaluates geographic language understanding across textual similarity and entity alignment tasks.

\paragraph{Spatial reasoning}
Spatial reasoning has received increasing attention alongside broader interest in general-purpose reasoning for LLMs. Early benchmarks focus on abstract, grid-based, or shape-centric reasoning. \citet{yamada2024evaluating} show that LLMs possess basic understanding of common shapes such as square but struggle with complex shapes such as hexagon.
SpatialEval~\citep{wang2024spatial} evaluates multimodal spatial understanding across navigation, grid reasoning, and map interpretation tasks, revealing substantial performance degradation as spatial complexity increases, while SpatialBench \citep{xu2025spatialbench} focuses on 3D and metric spatial reasoning in vision-language models, exposing persistent failures in depth, proximity, and relative positioning.
More aligned with geographical reasoning, GeoGramBench \citep{luo2025geogrambench} tests translation of geometry code to spatial representations, with frontier models achieving $<$50\% accuracy over problems of the higher abstraction levels. MapEval \citep{dihan2025mapeval} assesses map-based reasoning across 700 questions, where no model surpasses 67\% on spatial relationships. 
For geoparsing, benchmarks such as LGL \citep{lieberman2010geotagging} and WikToR \citep{gritta2018wikor} evaluate text-to-coordinate mapping.

\paragraph{Tool-use}
A separate line of work evaluates LLMs’ ability to orchestrate external Geographic Information System (GIS) tools. GeoBenchX \citep{krechetova2025geobenchx} benchmarks over 200 multi-step tasks involving 23 GIS tools, while GeoAnalystBench \citep{zhang2025geoanalystbench} evaluates Python code generation for spatial analysis workflows. Systems such as LLM-Geo \citep{li2023llmgeo} and GeoGPT~\citep{zhang2024geogpt} demonstrate end-to-end geospatial problem solving through tool integration.

\dsName\ differs from existing benchmarks in three key aspects. First, unlike geographic knowledge benchmarks that test factual recall (e.g., country statistics or place attributes), \dsName\ requires \textit{coordinate-level reasoning}, computing distances, bearings, and spatial relationships from raw GPS coordinates. Second, unlike tool-augmented benchmarks that evaluate API orchestration, \dsName\ tests \textit{intrinsic capabilities} encoded in model parameters, revealing training gaps that external tools would mask. Third, unlike mathematical reasoning benchmarks focused on planar geometry, \dsName\ uniquely tests \textit{geodetic computation}, e.g.\ Haversine distances, great-circle interpolation, spherical polygon areas, formulae essential for geographic applications but absent from existing benchmarks.


\section{\dsName}
\label{sec:gpsbench}

\dsName\ comprises \textbf{57,800 samples} across 17 tasks, organized into two tracks: the \textit{Pure GPS Track} (9 tasks) tests coordinate operations, and the \textit{Applied Track} (8 tasks) tests geographic knowledge. Each task contains 3,400 samples split into train (60\%), dev (10\%), and test (30\%) sets.

\subsection{Benchmark Design}
\label{sec:benchmark}

\dsName\ evaluates intrinsic geospatial knowledge, without relying on any external tools or resources. This distinction matters because: 
(1) we intend to test models' internal representations of geospatial knowledge to reveal patterns and biases that otherwise would be masked through tool-use; and (2) many use-cases require low latency, offline access, or privacy that invalidate tool use.

Following the landmark-route-survey framework from spatial cognition research \citep{siegel1975development}, we organize tasks by the type of spatial representation required:

\begin{itemize}[leftmargin=*, itemsep=2pt]
    \item \textbf{Landmark (L):} Recognizing individual locations from coordinates, the most basic spatial representation, requiring no metric relations between places.
    \item \textbf{Route (R):} Sequential, procedural understanding of paths connecting locations, egocentric representations learned through navigation.
    \item \textbf{Survey (S):} Allocentric, map-like representations with metric information enabling global spatial reasoning across locations.
    \item \textbf{Geometric (G):} Pure mathematical operations on coordinates requiring no geographic knowledge, a control condition isolating computational from world-knowledge capabilities.
\end{itemize}

Table~\ref{tab:task_summary} summarizes the  tasks included with their spatial knowledge types.

\begin{table*}[t]
\caption{Summary of \dsName\ tasks organized by spatial knowledge type following the landmark-route-survey framework. Types: \textbf{L}=Landmark, \textbf{R}=Route, \textbf{S}=Survey, \textbf{G}=Geometric. Samples are shorten for brevity.}
\label{tab:task_summary}

\centering
\begin{tabular}{@{}llp{9cm}c@{}}
\toprule
\textbf{Track} & \textbf{Task} & \textbf{Sample} & \textbf{Type} \\
\midrule
\multirow{9}{*}{\rotatebox[origin=c]{90}{Pure GPS}}
& Format Conversion & Convert 32°3'9.0"S, 115°53'16.2"E to decimal $\rightarrow$ \textit{-32.0525, 115.8878} & G  \\
& Coordinate Transformation & Transform (19.37, 95.22) to Universal Transverse Mercator (UTM) $\rightarrow$ \textit{Zone 46N, 732668E, 2143080N} & G  \\
& Distance Calculation & Distance from (-2.13, -47.56) to (-22.63, -47.05)? $\rightarrow$ \textit{2,280 km} & G, S  \\
& Bearing Computation & Bearing from (49.28, -123.13) to (5.89, 5.68)? $\rightarrow$ \textit{55.2° (NE)} & G, S  \\
&  Coordinate Interpolation & 50\% along (29.93, 117.95) to (-6.99, 106.55)? $\rightarrow$ \textit{(11.53, 111.86)} & G, R  \\
& Area \& Perimeter & Area of polygon [(-2.85, 33.08), (21.15, 72.96), (29.35, 105.89)]? $\rightarrow$ \textit{30,829 km²} & G, S  \\
& Bounding Box & Center of [(-2.85, 33.08), (21.15, 72.96), (53.60, 24.75), ...]? $\rightarrow$ \textit{(7.82, 52.08)} & G, S  \\
& Route Geometry & Simplify 12-point path using the Ramer–Douglas–Peucker algorithm ($\epsilon$=500m) $\rightarrow$ \textit{keep [0,1,4,5,6,7,11]} & G, R  \\
& Relative Position & Direction from (40.25, -8.39) to (0.06, 34.29)? $\rightarrow$ \textit{West} & G, S  \\
\midrule
\multirow{8}{*}{\rotatebox[origin=c]{90}{Applied}}
& Place Association & City at (41.12, -8.65)? $\rightarrow$ \textit{Canidelo, Vila Nova de Gaia, Portugal} & L  \\
& Name Disambiguation & 4 cities named ``Nawābganj''. Which at (26.93, 81.20)? $\rightarrow$ \textit{India} & L  \\
& Relative Position & Mikkeli to Sosnovka? $\rightarrow$ \textit{South} & L, S  \\
& Proximity & Closest to Titirangi: Elsdorf, Itatim, Murray, Cianjur? $\rightarrow$ \textit{Murray} & L, S  \\
& Route Analysis & Canberra on Gröbenzell→Dois Irmãos? $\rightarrow$ \textit{No} & L, R  \\
& Spatial Patterns & Outlier (by distance): Mornington, Ringwood, Caroline Springs, Ballajura, Sonneberg? $\rightarrow$ \textit{Sonneberg} & L, S  \\
& Boundary Analysis & Group by continent: Corroios, Wimbledon, Toms River, Santa Ana $\rightarrow$ \textit{EU:
          Corroios, Wimbledon, NA: Toms River, Santa Ana} & L, S  \\
& Terrain Classification & Terrain at (-27.47, 153.03)? $\rightarrow$ \textit{Coastal} & L \\
\bottomrule
\end{tabular}
\end{table*}

\subsection{Data Generation}

\begin{figure}[htbp]
\centering
\includegraphics[width=\columnwidth]{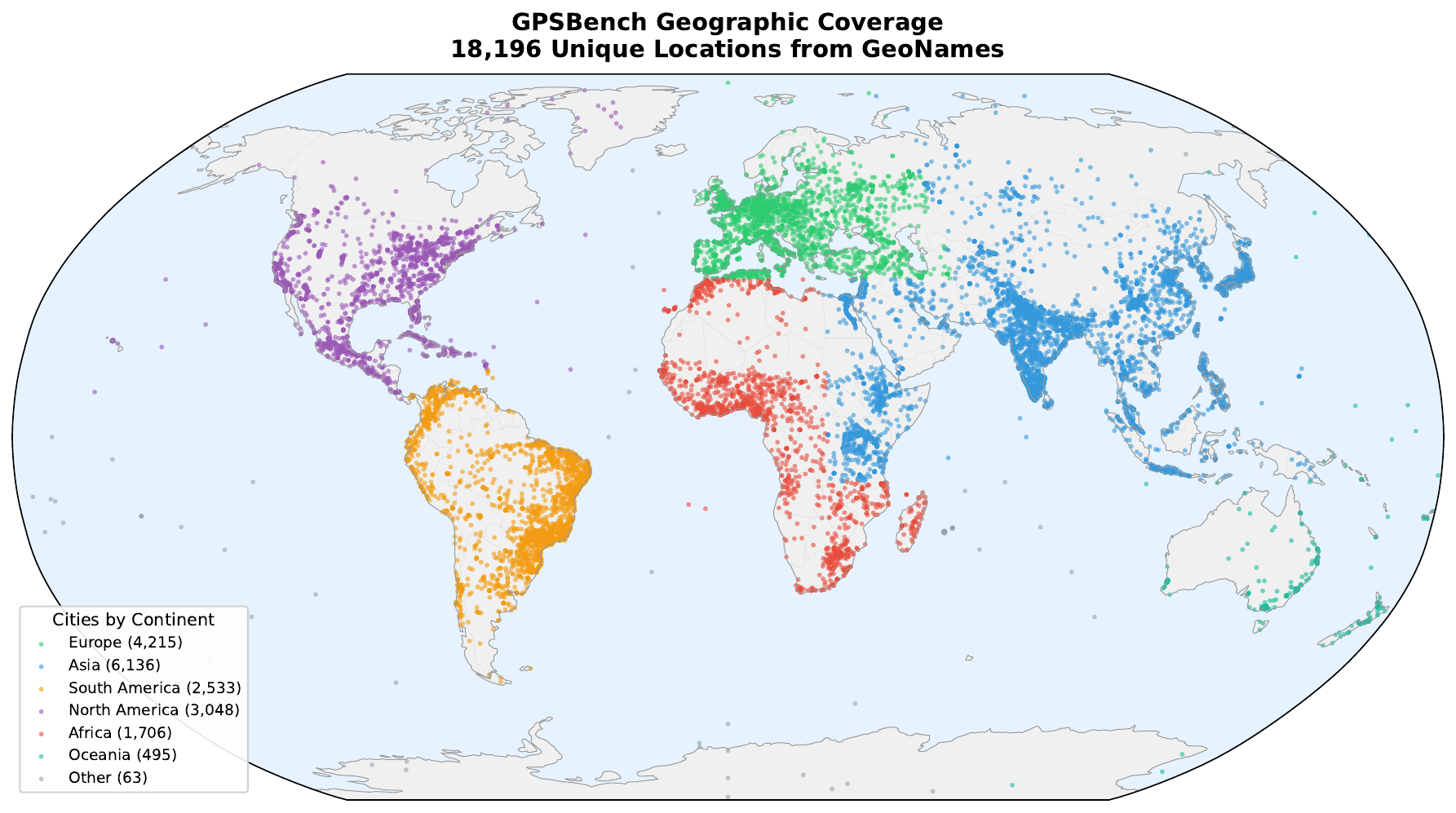}
\caption{Geographic coverage of \dsName. 18,196 unique locations from GeoNames span six  continents: Asia (33.7\%), Europe (23.2\%), North America (16.8\%), South America (13.9\%), Africa (9.4\%), and Oceania (2.7\%).}

\label{fig:world_map}
\end{figure}

\paragraph{Data source}
We leverage the \textbf{GeoName} database~\citep{GeoNames2026}, containing 32,709 cities with populations exceeding 15,000. Each entry provides WGS84 coordinates, city/country identifiers, population statistics, and alternate name spellings (3--5 variants per city), ensuring global coverage across 200+ countries. Figure~\ref{fig:world_map} shows the geographic distribution of 18,196 unique locations used in \dsName. Our location sampling strategies are detailed in Appendix~\ref{app:gpsbench_details}.

\paragraph{Sample generation pipeline}
Each task follows a structured generation pipeline: (1) city selection using task-appropriate sampling strategy (see Appendix~\ref{app:gpsbench_details}), (2) input construction by applying task-specific transformations, and (3) ground truth computation using geodetic formulae or database lookups. For Pure GPS tasks, we select cities globally, extract coordinates, and compute answers programmatically (e.g., Haversine distance or spherical bearing). For Applied tasks, we select cities based on task constraints (e.g., same-continent pairs for Relative Position, cities with duplicate names for Name Disambiguation), then test whether models can recover geographic facts from coordinates alone. All generation uses a fixed random seed for reproducibility. See Appendix~\ref{app:data_generation} for per-task generation details.

\paragraph{Ground truth computation}
All ground-truth values use standard geodetic formulae on the WGS84 ellipsoid approximated as a sphere ($R = 6371$~km): Haversine for distance, forward azimuth for bearing, spherical linear interpolation for intermediate points, and L'Huilier's theorem for polygon area. Coordinate transformations use Universal Transverse Mercator (UTM) and Web Mercator (EPSG:3857). The full formulae are included in Appendix~\ref{app:ground_truth_formulas}. Other non-computation tasks' ground truth are derived from GeoNames database, such as city names at sampled coordinates, country/continent metadata.

\section{Experiments}
\label{sec:experiments}

\subsection{Experiment setting}

\paragraph{Models evaluated} We evaluate 14 state-of-the-art LLMs spanning both proprietary and open-weight families: GPT~(GPT-5.1, GPT-5-mini, GPT-5-nano, GPT-4.1, GPT-4.1-mini)~\citep{openai2025gpt5, openai2025gpt41}, Gemini~(Gemini-2.5-Flash, Gemini-2.5-Pro)~\citep{team2025gemini}, Claude~(Claude-4.5-Haiku)~\cite{Claude}, Qwen3 (235B, 30B, 14B, and 8B)~\citep{yang2025qwen3}, and Mistral 2 (Small and Large)~\citep{mistral_large_2407}. Appendix~\ref{app:model_details} provides additional details of these models. 

\paragraph{Prompt format} All models are evaluated in a zero-shot setting using standardized prompts. The system prompt provides task-level context, while the user prompt contains the specific query. To ensure fair and comparable evaluation across models, we do not use chain-of-thought prompting or few-shot examples. Task samples and full prompt templates for the tasks are provided in Appendix~\ref{app:prompts}. 

\paragraph{Metrics} For multiple-choice tasks, we report standard accuracy. For numerical computation tasks, we compute the Mean Absolute Percentage Error (MAPE) between predicted and ground-truth values and report its complement, $1 - \text{MAPE}$, to align the directionality of scores with accuracy-based metrics, such that higher values consistently indicate better performance. This normalization allows direct aggregation and comparison across task types.

\subsection{Main Results}
\label{sec:main_results}

We evaluate 14 state-of-the-art LLMs on \dsName\ across two tracks: the Applied Track (8 tasks testing real-world geographic reasoning) and the Pure GPS Track (9 tasks testing coordinate manipulation).

\subsubsection{RQ1: How Well Do LLMs Perform on GPS Reasoning Tasks?}
\label{sec:rq1_overall_performance}

\begin{figure}[htbp]
\centering
\includegraphics[width=\columnwidth]{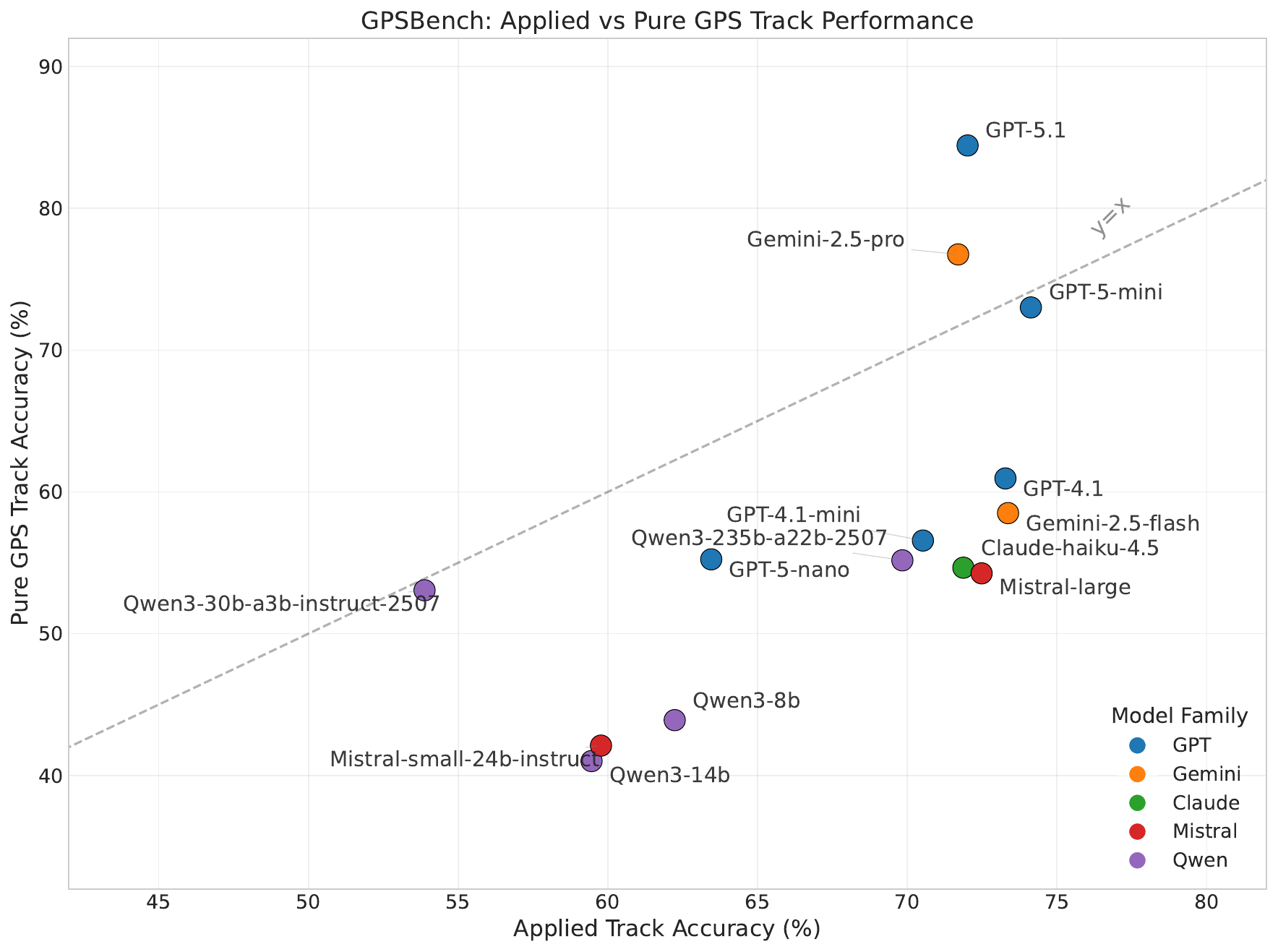}
\caption{Applied vs Pure GPS Track performance. Most models cluster below the diagonal (favoring Applied), while GPT-5.1 and Gemini-2.5-Pro uniquely excel at Pure GPS computation.}
\label{fig:scatter}
\end{figure}

\paragraph{Overall performance}
As shown in Figure~\ref{fig:scatter}, GPT-5-mini achieves the highest Applied Track accuracy (74.1\%), followed by Gemini-2.5-Flash (73.4\%) and GPT-4.1 (73.3\%). On the Pure GPS Track, GPT-5.1 leads with 84.4\%, followed by Gemini-2.5-Pro (76.7\%). The overall mean across 14 models is 67.7\% for Applied and 57.8\% for Pure GPS, with a +9.9\% gap favoring Applied tasks (see Appendix~\ref{app:overall_results} for full results).

\paragraph{Model family patterns}
Model families exhibit distinct capability profiles. GPT and Gemini models exhibit relatively balanced performance, with small Applied-Pure gaps (4--5\%). In contrast, Mistral, Claude, and Qwen models show substantially larger gaps (13--18\%), performing well on world-knowledge-driven tasks but struggling with coordinate-level computation. Notably, only two models exceed the diagonal in Figure~\ref{fig:scatter}: GPT-5.1 (Pure GPS $>$ Applied by 12.4\%) and Gemini-2.5-Pro (by 5.0\%). This pattern suggests that strong Pure GPS reasoning is not ubiquitous, but concentrated in flagship models with advanced mathematical capabilities.

GPT-5.1 achieves 84.4\% Pure GPS accuracy, surpassing GPT-5-mini (73.0\%) by 11.4\%. It dominates 5 of 9 Pure GPS tasks: Distance (99.9\%), Bearing (99.9\%), Route Geometry (96.3\%), Coordinate Transformation (90.8\%), and Interpolation (71.4\%). In contrast, Gemini-2.5-Pro leads on Applied tasks requiring world knowledge: Spatial Patterns (95.4\%) and Place Association (23.0\%, the highest among all models). This  indicates that GPS reasoning decomposes into two partially independent capabilities. See Appendix~\ref{app:track_gap} for additional discussion on performance patterns of different model families. 

\begin{figure}[htbp]
\centering
\includegraphics[width=\columnwidth]{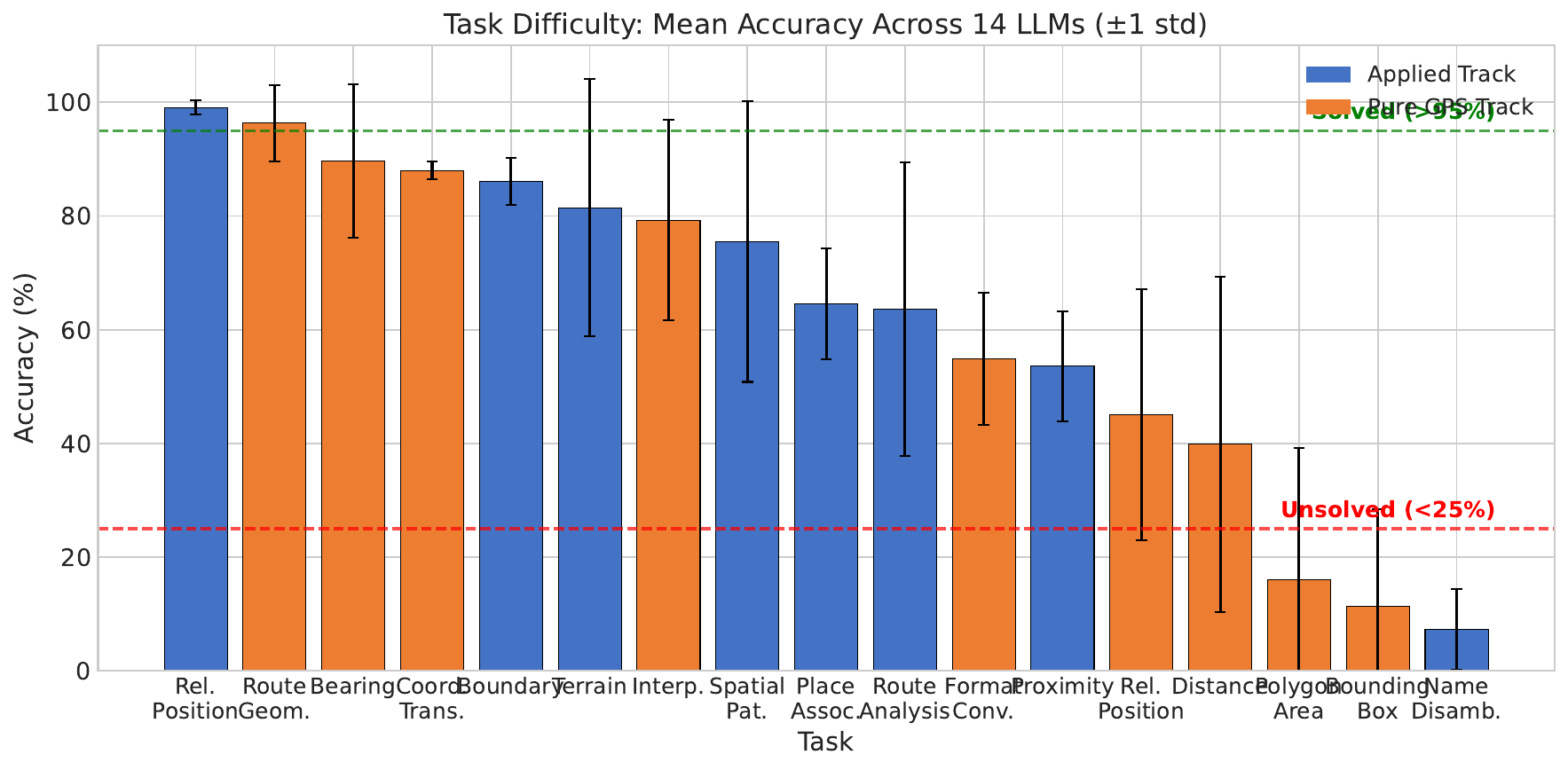}
\caption{Mean accuracy across all models per task, sorted by difficulty. Error bars show $\pm$1 standard deviation. Tasks cluster into solved ($>$95\%), brittle (25--95\%), and unsolved ($<$25\%) tiers.}
\label{fig:task_difficulty}
\end{figure}

\paragraph{Task difficulty}
Figure~\ref{fig:task_difficulty} reveals that task difficulty correlates with knowledge type rather than computational complexity. Tasks cluster into three tiers. \textit{Solved tasks} ($>$95\%) include Name Disambiguation and Bounding Box, which can be addressed through simple heuristics such as coarse spatial cues. \textit{Brittle tasks} (25--95\%) such as Distance Calculation, Coordinate Transformation, and Spatial Patterns show high variance across models, indicating  that these capabilities are unevenly distributed; some models excel while others fail entirely. \textit{Unsolved tasks} ($<$25\%) include Place Association, Polygon Area, and Coordinate Interpolation. Place Association requires dense coordinate-to-city mappings absent from training data, while Polygon Area and Interpolation demand multi-step spherical geometry reasoning. Notably, Polygon Area fails despite being purely mathematical, revealing that models have not learned geodetic formulae (see Appendix~\ref{app:per_task} for per-model breakdown). Qualitative analysis of model outputs indicates that failures arise from two primary sources: knowledge gaps (e.g., incorrect coordinate-to-place mappings, often defaulting to globally prominent cities within the correct country or region), computational limitations (e.g., incorrect formulae or error accumulation in multi-step calculations), and geographic bias, where models default to well-known locations when city names are ambiguous (e.g., assuming ``Rye'' refers to the English town rather than Rye, Australia, causing the model to misidentify the geographic outlier without doing the actual calculation). Representative error cases are discussed in Appendix~\ref{app:error_analysis}.

\begin{figure}[htbp]
\centering
\includegraphics[width=\columnwidth]{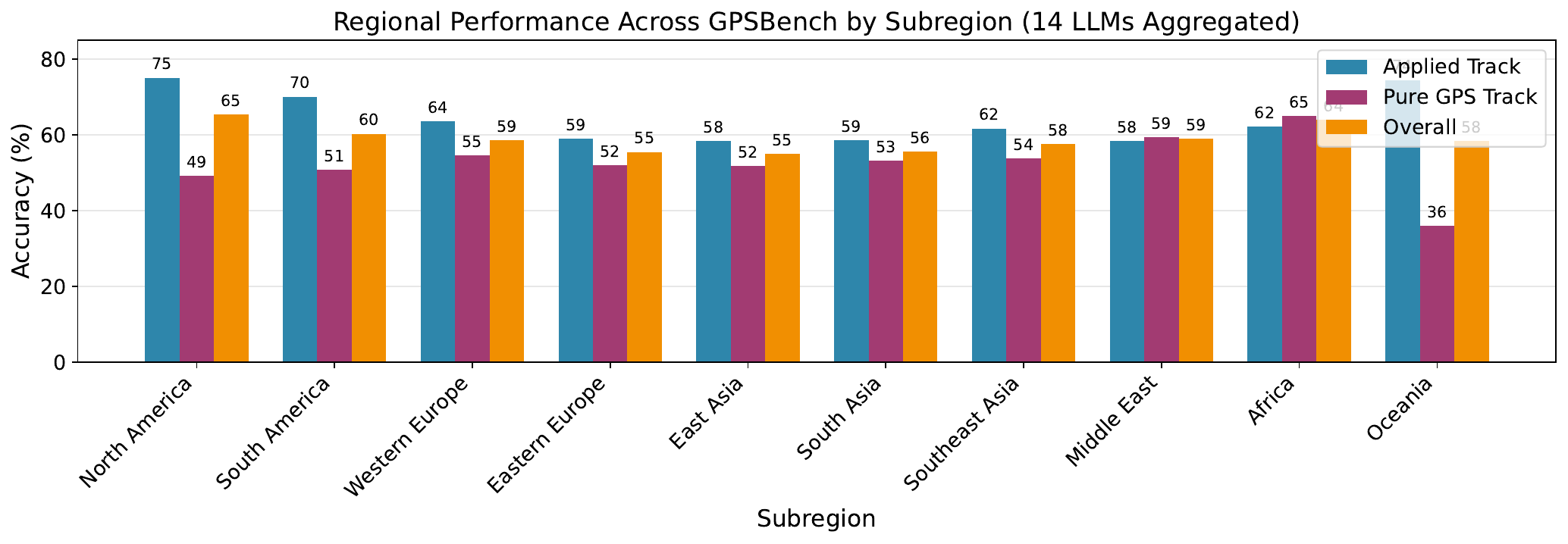}
\caption{Regional performance by subregion across all \dsName\ tasks. Applied Track shows the largest gaps: North America (75.8\%) vs.\ East Asia (59.6\%), a 16.2\% difference. Pure GPS Track is more uniform across most subregions.}
\label{fig:regional_subregion}
\end{figure}

\paragraph{Regional bias}
We further analyze performance across geographic subregions (Figure~\ref{fig:regional_subregion}). Averaged across all 17 tasks, overall accuracy varies by 10.6\% between subregions, from 66.4\% in North America to 55.8\% in East Asia. The Applied Track exhibits substantially larger regional variation (16.2\%), with North America (75.8\%) and Oceania (74.6\%) outperforming East Asia (59.6\%) and the Middle East (61.8\%). For Place Association in particular, disparities are stark: North America (14.7\%) and Western Europe (11.0\%) outperform South Asia (2.6\%) and East Asia (3.8\%) by factors of 4--6$\times$.

In contrast, Pure GPS computation tasks exhibit minimal regional bias. Performance on Distance Calculation (74.6--83.3\%), Bounding Box (95.2--98.5\%), and Format Conversion (97.4--100\%) varies by less than 9\% across all subregions. This divergence confirms that geographic bias primarily arises from uneven coordinate-to-place representations in training data rather than from limitations in geometric computation. This implies that location-aware applications that rely on world knowledge are likely to underperform in underrepresented regions, whereas purely geometric GPS reasoning remains comparatively robust worldwide. Detailed regional results are provided in Appendix~\ref{app:regional_performance}.

\subsubsection{RQ2: How Does Geographic Granularity Affect Performance?}
\label{sec:rq2_granularity_robustness}

We examine how model performance varies across levels of geographic granularity and assess robustness to coordinate noise, using the Place Association task as a diagnostic probe.

\begin{figure}[htbp]
\centering
\includegraphics[width=\columnwidth]{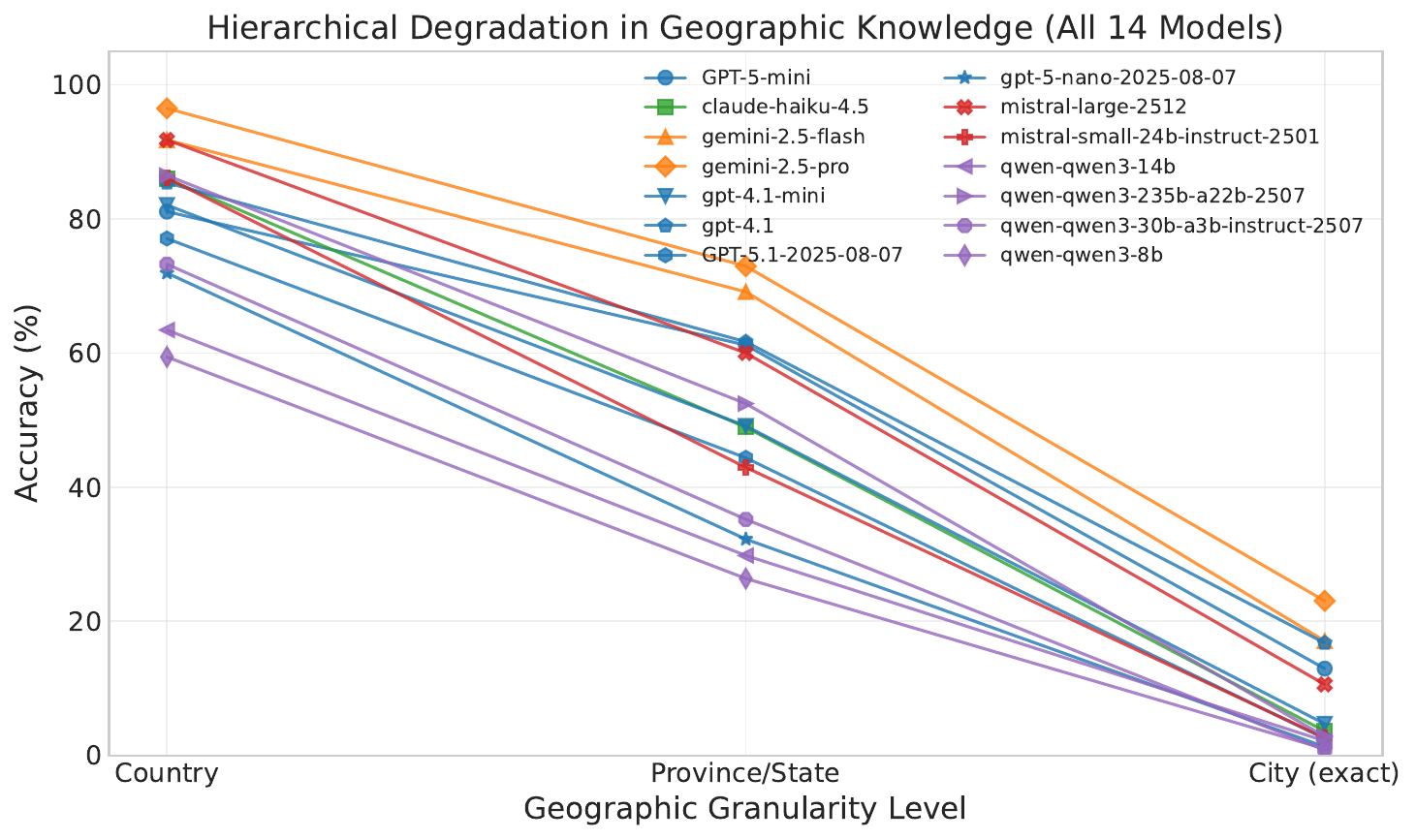}
\caption{Hierarchical degradation in geographic knowledge. Models perform well at country-level but accuracy drops sharply for finer granularities, with city-level accuracy below 25\% for all.}
\label{fig:place_assoc_granularity}
\end{figure}

\paragraph{Hierarchical degradation}
Figure~\ref{fig:place_assoc_granularity} reveals a clear hierarchy in geographic knowledge. Country-level identification accuracy ranges from 59--97\%, province/state-level accuracy drops to 26--73\%, and exact city identification collapses to 1--23\%. Together with the strong performance on the Name Disambiguation task, these results indicate that LLMs encode geographic knowledge primarily at coarse spatial resolutions, but lack the dense coordinate-to-city mappings required for fine-grained localization (cf.~Appendix~\ref{app:granularity_full}).

\paragraph{Robustness to noise}
To test sensitivity to coordinate precision and potential memorization, we perturb input coordinates with Gaussian noise ($\sigma \in \{0, 10, 50, 100, 500, 1000\}$ m), spanning conditions from high-precision GPS ($\pm$10 m) to coarse IP-based geolocation ($\pm$1 km). Each noise level contains approximately 567 samples. If models relied on memorized coordinate-place pairs, such perturbations would disrupt these mappings and degrade accuracy. We analyze performance at all granularity levels (Figure~\ref{fig:noise_granularity}, see detailed results in Appendix~\ref{app:noise_analysis}).

Across all noise levels, accuracy remains relatively stable. Country-level accuracy stays within 79--82\% ($\Delta = \pm 1.6\%$), province-level within 46--52\% ($\Delta = \pm 5.8\%$), and city-level within 6--9\% ($\Delta = \pm 2.0\%$). The stable performance  indicates that models rely on generalized geographic representations rather than memorized coordinate strings.

Further analysis of the 92.7\% of predictions that fail at city-level identification reveals that 80.4\% of these cases still correctly identify the country, while only 19.6\% miss the country entirely. This pattern is consistent across models: for example, Qwen3-8B achieves only 1.0\% city-level accuracy yet correctly identifies the country in 59.4\% of cases, whereas Gemini-2.5-Pro has 23.0\% city accuracy alongside 96.8\% country accuracy (23.0\% city-level and 73.8\% country-only). These results confirm that LLMs possess robust coarse-grained geographic knowledge but lack the dense coordinate-to-city mappings for precise localization.

To further test whether models have memorized entries from geographic databases such as GeoNames, we introduce a \emph{Missing Data} probe that requires inferring one coordinate (latitude or longitude) given a city name and the other coordinate (e.g., ``Given latitude 10.33384\textdegree and the location name San Diego, what is the longitude?''). If models had memorized GeoNames-style records, they would achieve high accuracy within a $\pm 0.1^\circ$ tolerance. Instead, performance is uniformly low across all models (mean: 8.3\%), with even the strongest model, Gemini-2.5-Pro, reaching only 12.4\%. This further supports the conclusion that LLMs do not memorize geocoding databases; rather, they encode coarse geographic structure while lacking fine-grained coordinate–place associations.

\begin{figure}[htbp]
\centering
\includegraphics[width=\columnwidth]{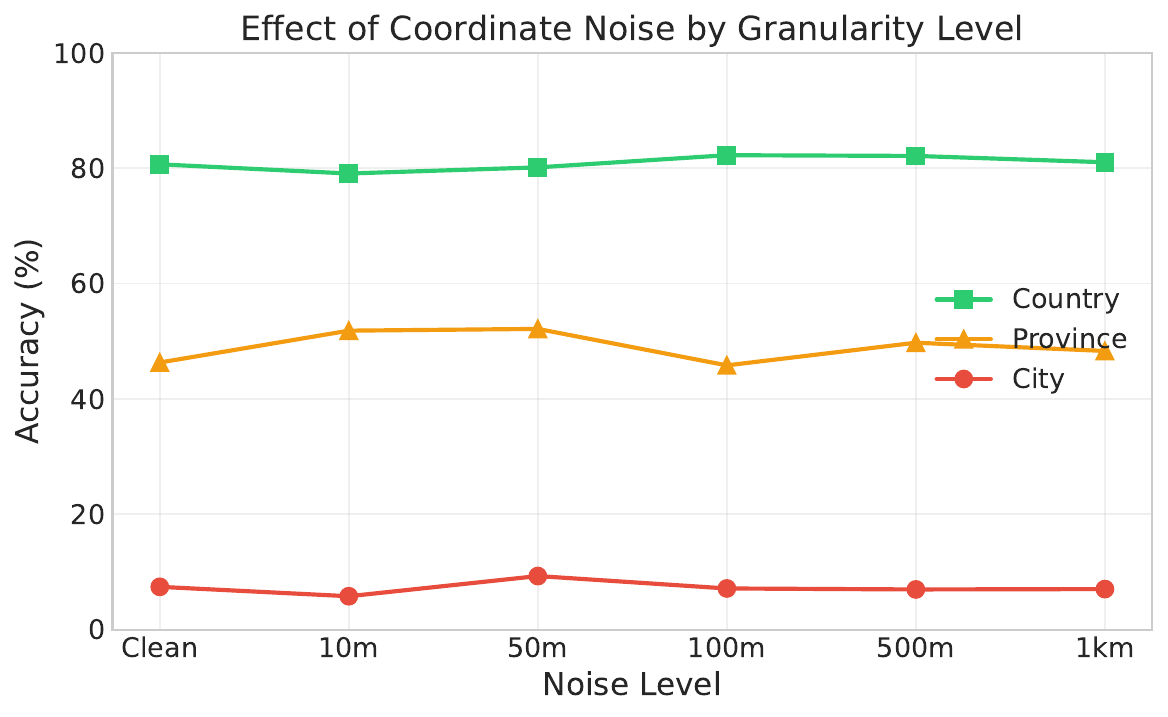}
\caption{Effect of coordinate noise across granularity levels. All levels show flat performance regardless of noise magnitude (10m to 1km), suggesting models rely on generalized geographic knowledge rather than memorized coordinate-to-place mappings.}
\label{fig:noise_granularity}
\end{figure}

\subsubsection{RQ3: Does GPS Augmentation Improve Downstream Geospatial Tasks?}
\label{sec:rq3_downstream_gps_augmentation}

Given that LLMs can process GPS coordinates for many tasks (RQ1), we investigate whether augmenting existing geographic reasoning benchmarks with explicit coordinates improves performance.

\paragraph{Datasets}
We evaluate on two benchmarks that test different aspects of geographic reasoning:
\begin{itemize}[leftmargin=*, itemsep=2pt]
    \item \textbf{MapEval}~\cite{dihan2025mapeval}: A benchmark of 200 map-based questions requiring spatial reasoning over real-world locations. Tasks include trip planning (``Which route is shorter from A to B?''), POI queries (``Find restaurants near the museum''), and nearby search. Questions reference specific place names without coordinates.
    \item \textbf{Hierarchical Spatial}~\cite{fulman2024distortions}: A diagnostic benchmark of 22 questions testing intercardinal direction judgments between city pairs (e.g., ``Is Toronto northeast or northwest of Portland?''). The benchmark is designed to expose systematic biases in spatial reasoning, including hierarchical bias (assuming relative positions based on country-level geography) and alignment bias (assuming cities align cardinally).
\end{itemize}

For MapEval, we geocode all place names mentioned in each question using the Google Maps API, then append a \texttt{[GPS Reference Coordinates]} section to the prompt listing latitude/longitude for each location (e.g., ``Cusco Cathedral: 13.5163°S, 71.9779°W''). We filter to samples where all places were successfully geocoded (n=66 of 200). For Hierarchical Spatial, we prepend each city's coordinates directly to the question (e.g., ``Portland, OR is located at coordinates (45.5202°N, 122.6742°W). Toronto, ON is located at coordinates (43.6532°N, 79.3832°W). Is Toronto northeast or northwest of Portland?'').

\begin{figure}[htbp]
\centering
\includegraphics[width=\columnwidth]{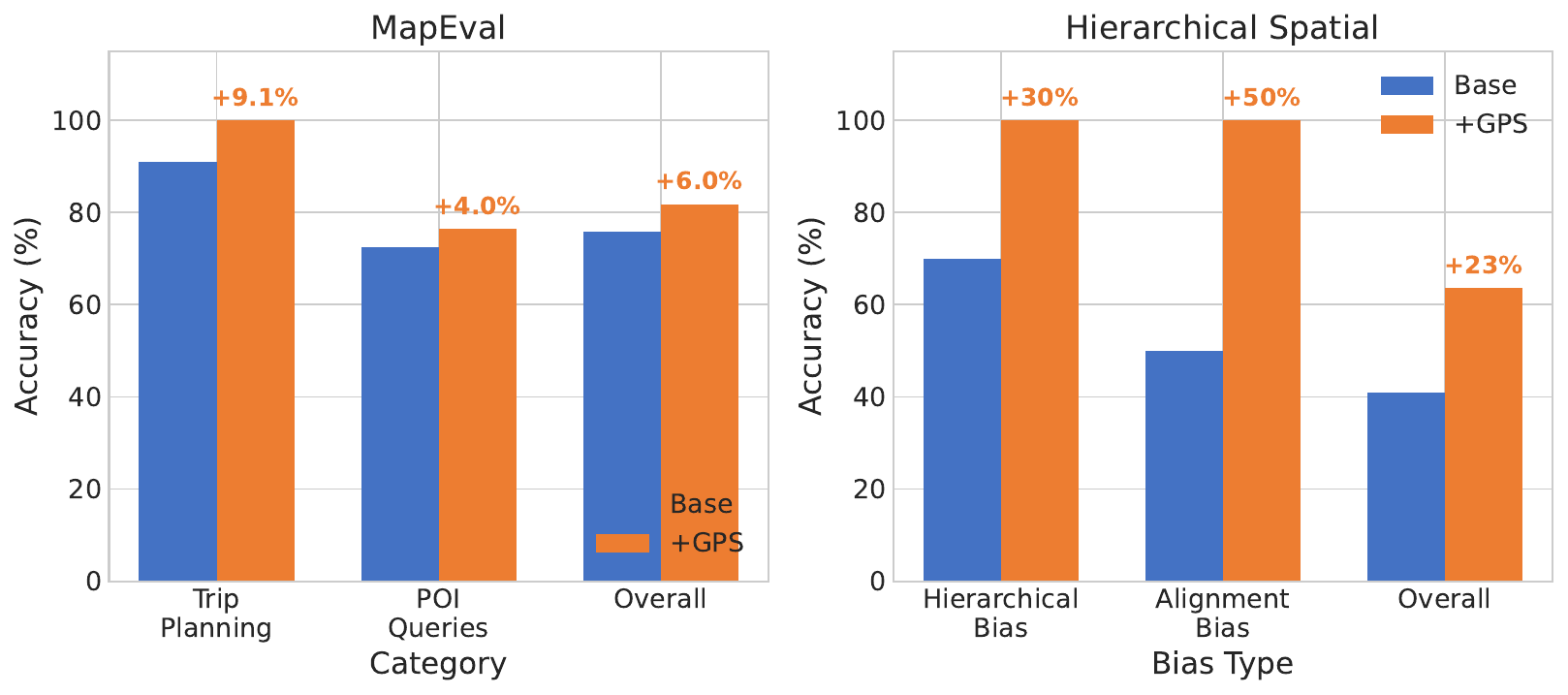}
\caption{Effect of GPS coordinate augmentation on downstream tasks. MapEval shows +6.1\% improvement; Hierarchical Spatial shows +22.7\%, with GPS eliminating hierarchical and alignment biases.}
\label{fig:downstream_gps}
\end{figure}

\paragraph{Findings}
GPS augmentation produces consistent improvements (Figure~\ref{fig:downstream_gps}). On MapEval, accuracy improves by +6.1\% (75.8\%→81.8\%) when all locations are geocoded. Trip planning benefits most (+9.1\%), as coordinates enable precise distance comparisons. On Hierarchical Spatial, GPS coordinates eliminate hierarchical bias (70\%→100\%) and alignment bias (50\%→100\%). For example, models previously assumed Toronto is north of Portland because Canada is north of the US; with coordinates, they correctly compute that Portland (45.5°N) is north of Toronto (43.7°N). However, proximity and rotation biases remain at 0\%, suggesting these errors stem from deeper spatial reasoning limitations rather than missing coordinate information. Overall, GPS augmentation provides +6--23\% benefit for tasks requiring precise spatial relationships (detailed in  Appendix~\ref{app:downstream}).

\subsubsection{RQ4: Can Finetuning Improve GPS Reasoning?}
\label{sec:rq4_finetuning}

We investigate whether task-specific finetuning can improve GPS reasoning by training Qwen3-30B on \dsName\  training data and comparing against its zero-shot baseline.

\begin{figure}[htbp]
\centering
\includegraphics[width=\columnwidth]{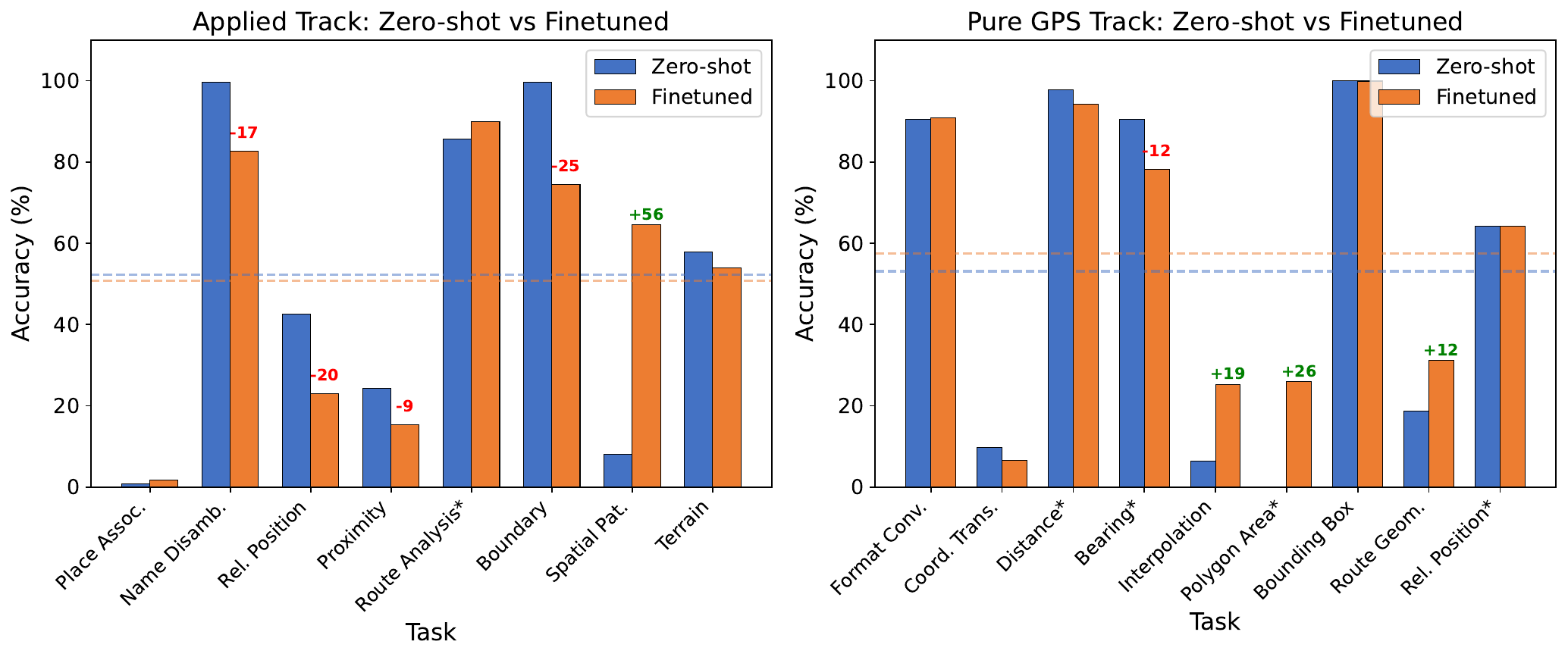}
\caption{Zero-shot vs.\ finetuned Qwen3-30B. Finetuning improves geometric computation but degrades world-knowledge tasks.}
\label{fig:finetuning}
\end{figure}

As shown in Figure~\ref{fig:finetuning}, finetuning yields mixed outcomes: Applied Track slightly degrades ($-$1.6\%) while Pure GPS improves (+4.3\%), with +1.5\% net improvement overall. At the task level, geometric reasoning improves substantially: Spatial Patterns (+56.5\%), Polygon Area (+25.9\%), Interpolation (+18.9\%), Route Geometry (+12.4; while world-knowledge tasks degrade: Boundary Analysis ($-$25.2\%), Name Disambiguation ($-$17.1\%). Notably, Bearing also degrades ($-$12.3\%) despite being a Pure GPS task, suggesting there is some degree of task-specific variability.

The pattern suggests finetuning strengthens coordinate computation at the expense of world knowledge integration. This highlights that GPS reasoning comprises two distinct capabilities: (1) coordinate manipulation, which is learnable, and (2) geographic knowledge, which risks degradation during finetuning. Future work should explore continual learning techniques to preserve base capabilities while adding GPS skills (see Appendix~\ref{app:finetuning} for per-task breakdown).

\subsubsection{RQ5: How Does Model Scale Affect GPS Reasoning?}
\label{sec:rq5_scaling}

We analyze scaling behavior using model families with known parameter counts: Mistral (24B and 123B) and Qwen3 (8B, 14B, 30B, and 235B).

\begin{figure}[htbp]
\centering
\includegraphics[width=\columnwidth]{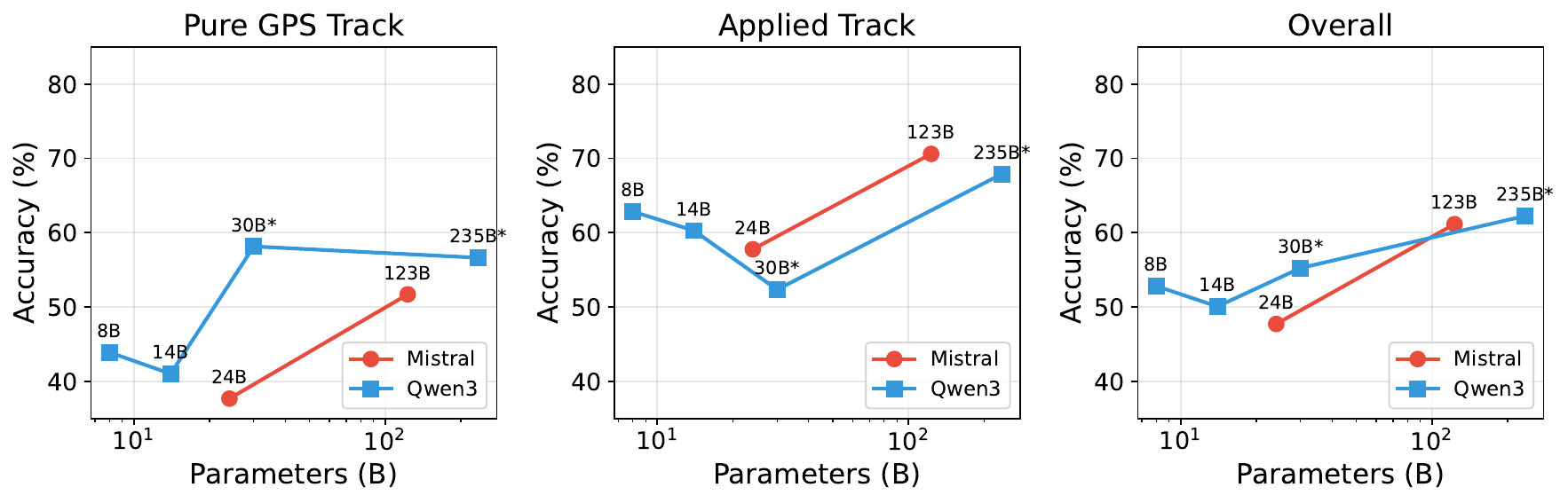}
\caption{Scaling analysis for Mistral and Qwen3 model families. Both families show overall improvement with scale.}
\label{fig:scaling}
\end{figure}

As shown in Figure~\ref{fig:scaling}, both model families improve with scale, though with different dynamics. Mistral shows consistent gains with 5.1$\times$ scale-up (24B$\rightarrow$123B). Qwen3 exhibits slight degradation at smaller sizes (8B$\rightarrow$14B) but strong improvement at larger scales. Across the full Qwen3 range (8B$\rightarrow$235B, 29$\times$), we observe substantial gains: Pure GPS +19.7\%, Applied +11.1\%, and Overall +16.0\%.
These patterns reveal that scaling remains an effective strategy to encode geographic knowledge and reasoning capabilities.


\section{Conclusion}
\label{sec:conclusion}

We introduced \dsName, a comprehensive benchmark for evaluating GPS reasoning in LLMs, spanning geometric coordinate computation and applied geographic reasoning tasks. Our evaluation reveals that GPS reasoning remains partially solved: models perform reasonably on basic geometric operations but struggle with complex spherical geometry and fine-grained place association.
We find that world knowledge does not transfer to coordinate computation, with most models showing stronger applied reasoning than pure GPS skills. Geographic bias is task-specific---applied tasks exhibit substantial regional disparities driven by sparse coordinate-to-place mappings, while pure computation remains region-agnostic.
Models encode coarse geography but lack fine-grained localization, and robustness to coordinate noise confirms genuine understanding rather than memorization. Finetuning reveals capability trade-offs where geometric computation improves at the expense of world knowledge.

\section{Limitations}
\label{sec:limitations}

\paragraph{Scope and generalization}
We evaluate only text-based LLMs; multimodal models that process maps, satellite imagery, or street-view photographs may exhibit different capability profiles. Our evaluation uses zero-shot prompting without chain-of-thought reasoning or tool augmentation, which may underestimate achievable performance with more sophisticated prompting strategies. All prompts are in English, leaving out cross-lingual geographic reasoning.

Geographic coverage is constrained by GeoNames data availability, which favors urban areas and may underrepresent rural areas. Population-based sampling means larger cities are overrepresented relative to their geographic footprint. Regions such as maritime, polar, and deserts are absent from our evaluation, as well as building-level geographic reasoning tasks.

\paragraph{Point-based city representation}
GeoNames provides a single reference coordinate per city, but cities are geographic areas spanning many square kilometers. A coordinate near a city boundary might reasonably map to multiple cities, yet our evaluation treats the GeoNames reference point as definitive. This is particularly problematic for Place Association, where a model answering with an adjacent city may have valid geographic reasoning but receives zero credit.

\paragraph{Distance-based error}
We use percentage-based error that treats all scales equivalently. A 5\% error on a 100 km distance (5 km) has different practical implications than 5\% on 10,000 km (500 km). Navigation may require meter-level precision locally but tolerate kilometer-level errors for continental distances.

\clearpage
\section*{Impact Statement} 

This paper presents work whose goal is to advance the evaluation of geographic and spatial reasoning in large language models. By systematically identifying where models succeed and fail across geographic granularity, our results can help guide the development of future LLMs and applications that rely on them, particularly by clarifying when coarse geographic reasoning is reliable and when fine-grained localization should not be trusted.
Our findings help reduce the risk of misapplication of LLMs in location-sensitive settings by making their limitations explicit, and can inform the design of safer, more robust downstream systems that incorporate validation or external tools when precise geographic reasoning is required. The benchmark uses only publicly available geographic data and does not involve personal or sensitive location information.

\bibliography{references}
\bibliographystyle{icml2026}

\newpage
\appendix

\section{\dsName\ Data Sampling Strategies}
\label{app:gpsbench_details}
\subsection{Sampling Strategies}
\label{app:sampling}
The GeoNames database exhibits natural geographic skew: Asia contains 37\% of cities, Europe 24\%, while Oceania has only 1\%. To prevent model performance from reflecting this imbalance, we implement quota-based sampling with target proportions: Africa: 20\%, Asia: 25\%, Europe: 20\%, North America: 15\%, South America: 15\%, Oceania: 5\%.


This rebalancing ensures models cannot exploit regional biases while still representing population distribution. All global samples enforce a minimum 100~km separation to avoid near-duplicate cities. For samples requiring fewer than 6 cities, we use weighted random selection from continents; for larger samples, we use proportional allocation with adjustments to match the exact target count.

\subsection{Regional Sampling}

For polygon-based tasks (Area \& Perimeter), we constrain vertex selection to the same continent within 50--2,000~km of an anchor city. This produces realistic sub-continental regions (country or state-sized) rather than globe-spanning shapes that would be geometrically trivial.

The algorithm proceeds as follows:
\begin{enumerate}[leftmargin=*, itemsep=1pt]
    \item Select a random continent with sufficient cities ($\geq 2n$ where $n$ is the required vertex count).
    \item Choose a random anchor city from that continent.
    \item Find all cities within 50--2,000~km of the anchor.
    \item Sample $n-1$ additional cities while enforcing minimum 50~km inter-vertex distance.
    \item Sort vertices by angle from centroid to ensure valid polygon ordering.
\end{enumerate}

\subsection{Same-Continent Sampling}

The Applied Track's Relative Position task uses same-continent city pairs with controlled distance ranges to ensure difficulty diversity:

\begin{itemize}[leftmargin=*, itemsep=1pt]
    \item 200--500~km (very close): Hardest, requires precise regional knowledge.
    \item 500--1,000~km (close): Difficult, same country or neighboring countries.
    \item 1,000--2,000~km (moderate): Medium, same general region.
    \item 2,000--3,000~km (far): Easier but still requires continental knowledge.
\end{itemize}

The generator cycles through these ranges to ensure balanced difficulty distribution. This makes questions substantially harder than those of the same task in the Pure GPS Track (which examine relative position of cross-continental cities; detailed next), comparing cities within the same country (e.g., ``Is Lyon east of Paris?'') requires actual geographic knowledge, whereas cross-continental comparisons (e.g., ``Is Tokyo east of Paris?'') are trivially answered by hemisphere.

\subsection{Difficulty Stratification}

Place Association samples are stratified by city population to control difficulty: 
\begin{itemize}[leftmargin=*, itemsep=1pt]
    \item \textbf{Easy} (population $\geq$ 500k): Major cities with distinctive locations that are well-known globally.
    \item \textbf{Medium} (100k--500k): Regional cities requiring area knowledge but still significant urban centers.
    \item \textbf{Hard} ($<$ 100k): Small cities in potentially densely populated regions where multiple candidates exist nearby.
\end{itemize}

\section{Task-specific Data Generation}
\label{app:data_generation}

This section provides detailed per-task sample generation procedures. All generation uses a fixed random seed (42) for reproducibility.

\subsection{Pure GPS Track Generation}

\paragraph{Format conversion} 
For each sample: (1) Select a random city from GeoNames. (2) Extract its decimal coordinates. (3) Randomly choose conversion direction, i.e., Decimal Degrees (DD) $\leftrightarrow$ Degrees-Minutes-Seconds (DMS). (4) Apply the conversion formula and store both representations. Ground truth is computed programmatically using standard conversion: degrees = floor(decimal), minutes = floor((decimal - degrees) $\times$ 60), seconds = remainder $\times$ 60.

\paragraph{Coordinate transformation}
For each sample: (1) Select a random city. (2) Randomly choose transformation type (WGS84 $\rightarrow$ UTM, WGS84 $\rightarrow$ Web Mercator, or reverse). (3) Compute the target coordinates using standard projection formulae. UTM zone is computed as $\lfloor(\lambda + 180)/6\rfloor + 1$. Tolerances: UTM $\pm$10m, Web Mercator $\pm$100m.

\paragraph{Distance calculation}
For each sample: (1) Select 2 cities using global balanced sampling with minimum 50~km separation. (2) Compute Haversine distance. (3) Store distance in both kilometers and miles. Tolerance: max(5~km, 5\% of distance).

\paragraph{Bearing computation}
For each sample: (1) Select 2 globally balanced cities with minimum 100~km separation. (2)~Compute initial bearing using spherical trigonometry. (3)~Map bearing to 8 cardinal directions (N, NE, E, SE, S, SW, W, NW) using 45° sectors. Tolerance: $\pm$5°.

\paragraph{Coordinate interpolation} 
For each sample: (1) Select 2 globally balanced cities with minimum 500~km separation. (2) Randomly choose interpolation fraction from \{0.25, 0.5, 0.75\}. (3) Compute intermediate point using spherical linear interpolation (slerp): convert to 3D unit vectors, interpolate along great circle arc, convert back to lat/lon. Tolerance: $\pm$0.01°.

\paragraph{Area \& perimeter} 
For each sample: (1) Select $n=$ 3--6 cities from the same continent using regional sampling (50--2,000~km range). (2) Sort vertices by angle from centroid to form a valid polygon. (3) Compute area using L'Huilier's theorem via triangulation from first vertex. (4) Compute perimeter as sum of Haversine distances between consecutive vertices. Tolerance: $\pm$5\%.

\paragraph{Bounding box}
For each sample: (1) Select 3--8 globally balanced cities. (2) Compute min/max latitude and longitude. (3) Calculate center as midpoint of bounds. (4)~Calculate centroid as arithmetic mean of all coordinates. (5)~Compute diagonal distance using Haversine.

\paragraph{Route geometry}
For each sample: (1) Generate a path of 8--15 waypoints along a route. (2) Apply Ramer–Douglas–Peucker  simplification with randomly chosen epsilon (0.5--2.0~km). (3) Store original points and indices of retained points after simplification.

\paragraph{Relative position (Pure GPS)}
For each sample: (1) Select 2 globally balanced cities with minimum 100~km separation. This ensures cities are typically on different continents, making direction obvious from hemisphere. (2) Compute bearing and map to cardinal direction. Note: This task is intentionally easier than that in the Applied Track.

\subsection{Applied Track Generation}

\paragraph{Place association}
For each sample: (1) Select a random city from GeoNames. (2) Apply Gaussian noise to coordinates with $\sigma$ randomly chosen from \{0, 10, 50, 100, 500, 1000\}~m, cycling through levels for balanced distribution ($\sim$17\% each). Noise is converted to degrees: $\sigma_\phi = \sigma_m / 111000$, $\sigma_\lambda = \sigma_m / (111000 \cdot \cos\phi)$. (3) Assign difficulty based on population: easy ($>$500K), medium (100K--500K), hard ($<$100K). Ground truth is the original city name and any GeoNames alternate spellings.

\paragraph{Name disambiguation} 
For each sample: (1) Identify city names that appear 4+ times globally in GeoNames (847 unique name groups, e.g., ``Springfield'' with 34 instances). (2) Select 4 candidate cities sharing the same name from different countries/regions. (3) Choose one as the answer and add Gaussian noise (100--500~m) to its coordinates to prevent exact coordinate matching as a shortcut. (4)~Present candidates as multiple choice options with country identifiers.

\paragraph{Relative position (Applied)}
For each sample: (1) Select 2 cities from the \textit{same continent} using regional sampling. (2) Enforce distance constraints by cycling through ranges: 200--500~km (hardest), 500--1,000~km, 1,000--2,000~km, 2,000--3,000~km. This creates genuinely challenging questions (e.g., ``Is Lyon east of Paris?'') unlike Pure GPS where different continents make direction trivial. (3) Present using city names only (no coordinates), requiring geographic knowledge.

\paragraph{Proximity}
For each sample: (1) Select 5 globally balanced cities with minimum 100~km separation. (2) Designate first city as reference point. (3) Compute Haversine distances to remaining 4 candidates. (4) Ground truth is the candidate with minimum distance. All distances are stored for evaluation.

\paragraph{Route analysis}
For each sample: (1) Select 3 globally balanced cities (start, end, waypoint) with minimum 200~km separation. (2) Compute direct great-circle distance (start $\rightarrow$ end). (3) Compute via-waypoint distance (start $\rightarrow$ waypoint $\rightarrow$ end). (4) Calculate detour percentage: $(d_{via} - d_{direct}) / d_{direct}$. (5) Label as ``on route'' if detour $<$10\%.

\paragraph{Spatial patterns (outlier detection)}
For each sample: (1)~Define 6 geographic regions (Europe, Asia, Africa, North America, South America, and Oceania). (2) Randomly select a main region and a different outlier region. (3) Sample 4 cities from main region and 1 from outlier region. (4)~Compute centroid and each city's distance to it. (5) Validate that exactly one city (the outlier) has distance $>2\times$ the median distance, with clarity ratio (furthest/second-furthest) $>$1.3 to ensure unambiguous answers.

\paragraph{Boundary analysis}
For each sample: (1) Select 4--6 globally balanced cities. (2) Retrieve country and continent from GeoNames metadata. (3) Create grouping task (e.g., ``Group by continent'') with ground truth derived from ISO 3166 country-to-continent mapping.

\paragraph{Terrain classification}
For each sample: (1) Select coordinates from curated geographic locations with unambiguous terrain types. (2) Categories: Urban (major city centers), Coastal (within 5~km of coastline), Mountain (elevation $>$2000 m), Desert (arid regions such as Sahara and Gobi), Forest (tropical/temperate forest regions), and Agricultural (major farming regions). (3) Labels validated against OpenStreetMap Nominatim and satellite imagery.

\subsection{Train/Dev/Test split}

Each task's 3,400 samples are split deterministically: Train (60\%, 2,040 samples), Dev (10\%, 340 samples), and Test (30\%, 1,020 samples). Splits are stratified to preserve difficulty distribution and geographic balance. Zero overlap is verified programmatically by checking sample IDs across splits.


\section{Ground Truth Computation}
\label{app:ground_truth_formulas}

All ground truth values use standard geodetic formulae on the WGS84 ellipsoid approximated as a sphere with radius $R = 6371$~km.

\subsection{Haversine Distance}

The great-circle distance between points $(\phi_1, \lambda_1)$ and $(\phi_2, \lambda_2)$:
\begin{equation}
d = 2R \arcsin\sqrt{\sin^2\!\left(\frac{\Delta\phi}{2}\right) + \cos\phi_1 \cos\phi_2 \sin^2\!\left(\frac{\Delta\lambda}{2}\right)}
\end{equation}
where $\Delta\phi = \phi_2 - \phi_1$ and $\Delta\lambda = \lambda_2 - \lambda_1$. This formula is accurate to within 0.5\% for most Earth distances.

\subsection{Initial Bearing}

The forward azimuth (initial bearing) from point $P_1$ to $P_2$:
\begin{equation}
\begin{split}
\theta = \text{atan2}\Big(&\sin\Delta\lambda \cos\phi_2, \\
&\cos\phi_1 \sin\phi_2 - \sin\phi_1 \cos\phi_2 \cos\Delta\lambda
\Big)
\end{split}
\end{equation}
The result is normalized to $[0°, 360°)$ and mapped to eight cardinal directions:
\begin{itemize}[leftmargin=*, itemsep=1pt]
    \item N: $[337.5°, 22.5°)$
    \item NE: $[22.5°, 67.5°)$
    \item E: $[67.5°, 112.5°)$
    \item SE: $[112.5°, 157.5°)$
    \item S: $[157.5°, 202.5°)$
    \item SW: $[202.5°, 247.5°)$
    \item W: $[247.5°, 292.5°)$
    \item NW: $[292.5°, 337.5°)$
\end{itemize}

\subsection{Great Circle Interpolation}

An intermediate point at fraction $f \in [0,1]$ along the great circle path is computed using spherical linear interpolation (slerp). First, convert coordinates to 3D unit vectors:
\begin{equation}
\mathbf{p} = (\cos\phi\cos\lambda,\, \cos\phi\sin\lambda,\, \sin\phi)
\end{equation}

Then interpolate:
\begin{align}
a &= \frac{\sin((1-f)\delta)}{\sin\delta}, \quad b = \frac{\sin(f\delta)}{\sin\delta} \\
\mathbf{p}_f &= a \cdot \mathbf{p}_1 + b \cdot \mathbf{p}_2
\end{align}
where $\delta$ is the angular distance between the points. Convert back to latitude and longitude:
\begin{equation}
\phi_f = \text{atan2}(z_f, \sqrt{x_f^2 + y_f^2}), \quad \lambda_f = \text{atan2}(y_f, x_f)
\end{equation}

\subsection{Spherical Polygon Area}
Spherical polygon area is calculated using L'Huilier's theorem via triangulation. For each spherical triangle with sides $a, b, c$ (angular distances) and semi-perimeter $s = (a+b+c)/2$, the spherical excess $E$ is:
\begin{equation}
\tan\!\left(\frac{E}{4}\right)
= \sqrt{\scriptstyle
\tan\!\left(\frac{s}{2}\right)
\tan\!\left(\frac{s-a}{2}\right)
\tan\!\left(\frac{s-b}{2}\right)
\tan\!\left(\frac{s-c}{2}\right)
}
\end{equation}

The total polygon area $A = R^2 \left|\sum_{i=1}^{n-2} E_i\right|$, where $n$ is the number of vertices.

\subsection{Coordinate Transformations}

\paragraph{UTM projection} 
The UTM projection divides Earth into 60 zones of 6° longitude each:
\begin{equation}
\text{Zone} = \left\lfloor\frac{\lambda + 180}{6}\right\rfloor + 1
\end{equation}

Easting and Northing are computed via Transverse Mercator projection with scale factor $k_0 = 0.9996$. The central meridian of each zone is:
\begin{equation}
\lambda_0 = (\text{Zone} - 1) \times 6° - 180° + 3°
\end{equation}

False easting of 500,000~m is added to ensure positive values. For southern hemisphere, false northing of 10,000,000~m is added.

\begin{table*}[h]
\centering
\caption{Model details for \dsName\ evaluation. All models use temperature 0 for reproducibility. Parameter counts are estimates where not officially disclosed.}
\label{tab:model_details}
\setlength{\tabcolsep}{2pt}
\small
\begin{tabular}{lllllll}
\toprule
\textbf{Model} & \textbf{Endpoint / Model ID} & \textbf{Params} & \textbf{Type} & \textbf{Temp.} & \textbf{Provider} & \textbf{API} \\
\midrule
GPT-5.1 & gpt-5.1 & undisclosed & Reasoning & 0 & OpenAI & OpenAI \\
GPT-4.1 & gpt-4.1 & undisclosed & Non-reasoning & 0 & OpenAI & OpenAI \\
GPT-5-mini & gpt-5-mini-2025-08-07 & undisclosed & Reasoning & 0 & OpenAI & OpenAI \\
GPT-4.1-mini & gpt-4.1-mini & undisclosed & Non-reasoning & 0 & OpenAI & OpenAI \\
GPT-5-nano & gpt-5-nano-2025-08-07 & undisclosed & Reasoning & 0 & OpenAI & OpenAI \\
\midrule
Gemini-2.5-Pro & gemini-2.5-pro & undisclosed & Reasoning & 0 & Google & Google AI \\
Gemini-2.5-Flash & gemini-2.5-flash & undisclosed & Non-reasoning & 0 & Google & Google AI \\
\midrule
Claude-Haiku-4.5 & anthropic/claude-haiku-4.5 & undisclosed & Non-reasoning & 0 & Anthropic & OpenRouter \\
\midrule
Qwen3-235B & qwen/qwen3-235b-a22b-2507 & 235B (22B active) & Reasoning (MoE) & 0 & Alibaba & OpenRouter \\
Qwen3-30B & qwen/qwen3-30b-a3b-instruct-2507 & 30B (3B active) & Reasoning (MoE) & 0 & Alibaba & OpenRouter \\
Qwen3-14B & qwen/qwen3-14b & 14B  & Reasoning (Dense) & 0 & Alibaba & OpenRouter \\
Qwen3-8B & qwen/qwen3-8b & 8B & Reasoning (Dense) & 0 & Alibaba & OpenRouter \\
\midrule
Mistral-Large & mistralai/mistral-large-2512 & $\sim$123B & Non-reasoning & 0 & Mistral AI & OpenRouter \\
Mistral-Small & mistralai/mistral-small-24b-instruct-2501 & 24B & Non-reasoning & 0 & Mistral AI & OpenRouter \\
\bottomrule
\end{tabular}
\end{table*}

\paragraph{Web Mercator projection (EPSG:3857)}
The Web Mercator projection used by web mapping services is:
\begin{equation}
x = R \lambda, \quad y = R \ln\!\left(\tan\!\left(\frac{\pi}{4} + \frac{\phi}{2}\right)\right)
\end{equation}
where $\lambda$ is in radians. This projection is valid for latitudes between approximately $\pm 85.05°$.

\subsection{Knowledge Task Ground Truth (Applied Track)}

Unlike geometric tasks with mathematical formulae, Applied Track tasks derive ground truth from authoritative geographic databases as follows. 

\paragraph{Place association}
Ground truth is the city name from GeoNames at the sampled coordinate. Evaluation accepts the canonical city name or any of its alternate spellings (3--5 variants per city in GeoNames).

\paragraph{Name disambiguation}
For cities with duplicate names globally (e.g., ``Springfield''), GeoNames provides distinct entries with unique coordinates. Ground truth is the option whose GeoNames coordinates match the query coordinates (within 500m tolerance to account for coordinate noise).

\paragraph{Relative position (Applied)}
Given two city names, we retrieve their GeoNames coordinates and compute the bearing using the formula in Section~\ref{app:ground_truth_formulas}. The bearing is mapped to 8 cardinal directions.

\paragraph{Proximity}
Ground truth is determined by computing Haversine distances from the reference city to all candidate cities and selecting the minimum. The correct answer is deterministic given the GeoNames coordinates.

\paragraph{Route analysis}
Given start, end, and query cities, we compute whether the query city lies within a corridor around the great-circle path. A city is ``on route'' if its perpendicular distance to the path is $<$ 500 km and it lies between the start/end latitudes (with tolerance).

\paragraph{Spatial patterns (outlier detection)}
For each set of 5 cities, we compute the centroid and measure each city's distance to it. The outlier is the city with the maximum distance, required to have a ratio $>1.3\times$ the second-farthest city to ensure unambiguous answers.

\paragraph{Boundary analysis}
Country and continent assignments come directly from GeoNames metadata fields (\texttt{country\_code}, mapped to continents via ISO 3166).

\paragraph{Terrain classification} 
Coordinates are sampled from well-known geographic locations with unambiguous terrain types (e.g., Brisbane coordinates for Urban, Sahara coordinates for Desert, and Himalayan coordinates for Mountain). Labels are validated against OpenStreetMap Nominatim and cross-referenced with satellite imagery and geographic databases.

\section{Model Details}
\label{app:model_details}

Table~\ref{tab:model_details} summarizes the 14 models evaluated.

\paragraph{Model selection} We selected models to cover diverse architectures (dense vs. MoE), scales (24B--235B+ parameters), and reasoning capabilities. All frontier model families with public API access as of January 2026 are represented.

\paragraph{Reasoning vs. non-reasoning} Models marked as ``Reasoning'' employ extended thinking or chain-of-thought capabilities. For GPT-5.x models, we set \texttt{reasoning\_effort=low} to minimize reasoning overhead. For Gemini Flash models, we set \texttt{thinking\_budget=0}. For OpenRouter models, we disabled reasoning via \texttt{reasoning.effort=none}. Non-reasoning models generate responses directly.

\paragraph{Inference settings} All evaluations use temperature 0 for deterministic outputs. Maximum output tokens were set to 8192 for all models to accommodate complex task responses. No system prompts were used beyond task-specific instructions (detailed in Appendix~\ref{app:prompts}).

\paragraph{API access} OpenAI models (GPT-4.1 and GPT-5.x) were accessed via the OpenAI API. Gemini models were accessed via Google AI Studio. All other models (Claude, Qwen, and Mistral) were accessed via OpenRouter.\footnote{\url{https://openrouter.ai/}}

\section{Prompt Templates and Examples}
\label{app:prompts}

We use the following prompt template in combined with the subsequent task-specific instructions to prompt the LLMs. 
\begin{systemprompt}
You are an expert in GPS coordinates, geographic information systems, and spatial reasoning.

Provide accurate, precise answers based on the given coordinates and geographic data.

You may show your reasoning or calculations, but you MUST always end with:
FINAL ANSWER: [your answer]

For multiple choice questions, include only the letter (A, B, C, D, etc.) in FINAL ANSWER.
For numeric answers, provide the number with appropriate units.
For location names, provide the specific location name.
\end{systemprompt}

\subsection{Pure GPS Track Tasks}

\textbf{Format Conversion}

\begin{taskprompt}\ttfamily\small
Convert the coordinate from DMS to decimal\_degrees.\\
Source coordinate (DMS): 43°55'31.8"N, 81°24'43.6"E\\
FINAL ANSWER: [converted coordinate]
\end{taskprompt}
\textit{Gold Answer:} 43.92550, 81.41211

\textbf{Distance Calculation}
\begin{taskprompt}\ttfamily\small
Calculate the distance between these two points:\\
Point A: Djenné (13.90608, -4.55332)\\
Point B: Parbhani (19.26855, 76.77081)\\
FINAL ANSWER: [distance] km
\end{taskprompt}
\textit{Gold Answer:} 8,610.5 km

\textbf{Bearing Computation}
\begin{taskprompt}\ttfamily\small
Calculate the initial bearing from Point A to Point B:\\
Point A: Tokyo (35.6762, 139.6503)\\
Point B: Sydney (-33.8688, 151.2093)\\
FINAL ANSWER: [bearing]°
\end{taskprompt}
\textit{Gold Answer:} 169.83\textdegree

\textbf{Coordinate Interpolation}
\begin{taskprompt}\ttfamily\small
Find the coordinates of the point that is 30\% of the way from Point A to Point B along the geodesic path:\\
Point A: London (51.5074, -0.1278)\\
Point B: New York (40.7128, -74.0060)\\
FINAL ANSWER: [latitude], [longitude]
\end{taskprompt}
\textit{Gold Answer:} 53.7722, -24.7404

\textbf{Coordinate System Transformation}
\begin{taskprompt}\ttfamily\small
Convert the following coordinates from WGS84 (EPSG:4326) to UTM Zone 46N:\\
\\
Location: Myaydo\\
Latitude: 19.36838\\
Longitude: 95.21512\\
\\
FINAL ANSWER: [converted coordinates]
\end{taskprompt}
\textit{Gold Answer:} 732668.82 E, 2143080.82 N (UTM Zone 46N)

\textbf{Polygon Area}
\begin{taskprompt}\ttfamily\small
Calculate the area of the polygon formed by these coordinates:\\
\\
~~Vertex 1: El Mansouria (33.74643, -7.30194)\\
~~Vertex 2: Beni Enzar (35.26, -2.93)\\
~~Vertex 3: Menzel Abderhaman (37.23737, 9.86313)\\
\\
Use spherical geometry (Earth radius $\approx$ 6371 km).\\
\\
FINAL ANSWER: [area] km\textsuperscript{2}
\end{taskprompt}
\textit{Gold Answer:} 30,829.07 km\textsuperscript{2}

\textbf{Bounding Box}
\begin{taskprompt}\ttfamily\small
Calculate the bounding box (minimum rectangle) containing these 7 points:\\
\\
~~- Misungwi (-2.85, 33.08333)\\
~~- Chalthan (21.15421, 72.96141)\\
~~- Yongchuan (29.35376, 105.89392)\\
~~- Shchuchyn (53.6014, 24.7465)\\
~~- Avon (41.80982, -72.83065)\\
~~- Coruripe (-10.12556, -36.17556)\\
~~- Whakatane (-37.95855, 176.98545)\\
\\
Provide the bounding box as: min\_lat, max\_lat, min\_lon, max\_lon\\
\\
FINAL ANSWER: [min\_lat], [max\_lat], [min\_lon], [max\_lon]
\end{taskprompt}
\textit{Gold Answer:} -37.95855, 53.6014, -72.83065, 176.98545

\textbf{Route Geometry}
\begin{taskprompt}\ttfamily\small
Analyze this route with 12 waypoints:\\
\\
~~Waypoint 1: (14.78333, -16.96667)\\
~~Waypoint 2: (14.770326, -16.98198)\\
~~Waypoint 3: (14.744601, -16.975185)\\
~~...\\
~~Waypoint 12: (14.15, -16.55)\\
\\
Calculate the total route length by summing distances between consecutive waypoints, then use it to answer the question.\\
\\
FINAL ANSWER: [your answer]
\end{taskprompt}
\textit{Gold Answer:} A (points kept: [0, 1, 4, 5, 6, 7, 11])

\textbf{Relative Position (Pure GPS)}
\begin{taskprompt}\ttfamily\small
What is the cardinal direction from Point A (0.0607, 34.28806) to Point B (40.24678, -8.39402)?\\
\\
A) North ~~B) South ~~C) East ~~D) West\\
\\
FINAL ANSWER: [letter]
\end{taskprompt}
\textit{Gold Answer:} D) West

\subsection{Applied Track Tasks}

\textbf{Place Association}
\begin{taskprompt}\ttfamily\small
What city is located at coordinates 48.8566, 2.3522?\\
FINAL ANSWER: [city name]
\end{taskprompt}
\textit{Gold Answer:} Paris, France

\textbf{Name Disambiguation}
\begin{taskprompt}\ttfamily\small
There are multiple cities named "Springfield". Which Springfield is located at coordinates 39.7817, -89.6501?\\
A) Springfield, Massachusetts\\
B) Springfield, Illinois\\
C) Springfield, Missouri\\
D) Springfield, Ohio\\
FINAL ANSWER: [letter]
\end{taskprompt}
\textit{Gold Answer:} B) Springfield, Illinois

\textbf{Proximity}
\begin{taskprompt}\ttfamily\small
Which of these cities is closest to Paris (48.8566, 2.3522)?\\
A) London (51.5074, -0.1278)\\
B) Berlin (52.5200, 13.4050)\\
C) Brussels (50.8503, 4.3517)\\
D) Amsterdam (52.3676, 4.9041)\\
FINAL ANSWER: [letter]
\end{taskprompt}
\textit{Gold Answer:} C) Brussels

\textbf{Relative Position (Applied)}
\begin{taskprompt}\ttfamily\small
What is the cardinal direction from Mikkeli to Sosnovka?\\
\\
A) North ~~B) South ~~C) East ~~D) West\\
\\
FINAL ANSWER: [letter]
\end{taskprompt}
\textit{Gold Answer:} B) South

\textbf{Route Analysis}
\begin{taskprompt}\ttfamily\small
Is Canberra approximately on the direct path from Gr\"{o}benzell to Dois Irm\~{a}os?\\
\\
FINAL ANSWER: [Yes/No]
\end{taskprompt}
\textit{Gold Answer:} No (detour: 170.4\%)

\textbf{Boundary Analysis}
\begin{taskprompt}\ttfamily\small
Group these cities by continent: Corroios, Wimbledon, Toms River, Santa Ana\\
\\
FINAL ANSWER: [grouping]
\end{taskprompt}
\textit{Gold Answer:} Europe: \{Corroios, Wimbledon\}, North America: \{Toms River, Santa Ana\}

\textbf{Spatial Patterns}
\begin{taskprompt}\ttfamily\small
Which location is the geographic outlier among: Gda\'{n}sk, Castlereagh, Vostryakovo, Shch\"{e}kino, Chiconcuac?\\
FINAL ANSWER: [city name]
\end{taskprompt}
\textit{Gold Answer:} Chiconcuac (located in Mexico; other four cities are in Europe)

\textbf{Terrain Classification}
\begin{taskprompt}\ttfamily\small
What type of terrain/environment is at coordinates -27.4698, 153.0251?\\
\\
A) Urban ~~B) Arctic/Ice ~~C) Ocean/Sea\\
D) Mountain ~~E) River/Lake ~~F) Desert\\
\\
FINAL ANSWER: [letter]
\end{taskprompt}
\textit{Gold Answer:} A) Urban (Brisbane, Australia)

\section{Additional Results}
\label{app:additional_results}

\subsection{Overall Performance Results}
\label{app:overall_results}

Table~\ref{tab:main_results_overall} presents   performance results of the different LLMs averaged over the tasks of each track.   

\begin{table}[htbp]
\caption{Overall model performance on \dsName\ (\% accuracy).}
\label{tab:main_results_overall}
\centering
\small
\begin{tabular}{lccc}
\toprule
\textbf{Model} & \textbf{Applied} & \textbf{Pure GPS} & \textbf{Gap} \\
\midrule
GPT-5.1 & 72.0 & \textbf{84.4} & $-$12.4 \\
GPT-5-mini & \textbf{74.1} & 73.0 & +1.1 \\
GPT-5-nano & 63.5 & 55.2 & +8.2 \\
GPT-4.1 & 73.3 & 60.9 & +12.3 \\
GPT-4.1-mini & 70.5 & 56.6 & +13.9 \\
\midrule
Gemini-2.5-Pro & 71.7 & 76.7 & $-$5.0 \\
Gemini-2.5-Flash & 73.4 & 58.5 & +14.9 \\
\midrule
Claude-Haiku-4.5 & 71.9 & 54.7 & +17.2 \\
\midrule
Mistral-Large & 72.5 & 54.3 & +18.2 \\
Mistral-Small-24B & 59.8 & 42.1 & +17.7 \\
\midrule
Qwen3-235B & 69.8 & 55.2 & +14.6 \\
Qwen3-30B & 53.9 & 53.1 & +0.8 \\
Qwen3-14B & 50.3 & 35.4 & +14.9 \\
Qwen3-8B & 47.1 & 33.3 & +13.8 \\
\bottomrule
\end{tabular}
\end{table}

\subsection{Applied vs.\ Pure GPS Gap Analysis}
\label{app:track_gap}


Table~\ref{tab:main_results_overall} reveals that 12 of 14 models score higher on the Applied Track than on the Pure GPS Track, with gaps ranging from +0.8\% (Qwen3-30B) to +18.2\% (Mistral-Large). The two exceptions are GPT-5.1 ($-$12.4\%) and Gemini-2.5-Pro ($-$5.0\%), both flagship reasoning models whose Pure GPS scores surpass their Applied scores. The mean gap across all 14 models is +9.9\%, confirming that world knowledge encoded during pretraining provides a substantial advantage over raw coordinate-level computation for the majority of current LLMs.

The GPT family (5 models) leads both tracks, averaging 70.7\% on Applied and 66.0\% on Pure GPS. A clear split emerges within this family: the reasoning models GPT-5.1 and GPT-5-mini show near-balanced or reversed gaps, while the non-reasoning GPT-4.1 and GPT-4.1-mini exhibit large positive gaps. The Gemini family (2 models) mirrors this pattern: Gemini-2.5-Pro (reasoning) achieves a $-$5.0\% gap while Gemini-2.5-Flash shows +14.9\%. The Mistral family (2 models) exhibits the largest average gap (+17.9\%), with both models showing nearly identical gaps (+18.2\% and +17.7\%) despite a 5$\times$ difference in parameter count, suggesting this reflects a training-data characteristic rather than a scale-dependent phenomenon. The Qwen family (4 models) provides the most granular scaling view: overall accuracy increases from 40.2\% (8B) to 62.5\% (235B), with the gap narrowing for MoE models. Notably, Qwen3-30B (3B active parameters) achieves a near-zero gap of +0.8\%, suggesting MoE architectures may allocate specialized experts to geometric computation.


\begin{table}[htbp]
\centering
\caption{Per-task accuracy (\%) over Applied Track tasks. Bold indicates best per task.}
\label{tab:per_task_applied}
\setlength{\tabcolsep}{1pt}
\resizebox{\linewidth}{!}{
\begin{tabular}{lcccccccc}
\toprule
\textbf{Model} & \textbf{Name} & \textbf{Route} & \textbf{Spatial} & \textbf{Bound.} & \textbf{Terrain} & \textbf{Prox.} & \textbf{Rel. Pos.} & \textbf{Place} \\
\midrule
GPT-5.1 & 97.7 & 81.7 & 93.7 & 98.1 & \textbf{72.6} & 73.3 & 58.5 & 2.5 \\
GPT-5-mini & \textbf{100.0} & 90.1 & 95.2 & 67.4 & 70.6 & \textbf{89.6} & 56.6 & 12.9 \\
GPT-5-nano & 99.9 & 84.8 & 59.4 & 98.5 & 67.4 & 33.1 & 52.6 & 1.3 \\
GPT-4.1 & \textbf{100.0} & 88.6 & 94.2 & 45.2 & 71.3 & 84.4 & 64.8 & 16.8 \\
GPT-4.1-mini & \textbf{100.0} & 89.3 & 92.5 & 56.0 & 70.4 & 82.4 & 53.1 & 4.7 \\
\midrule
Gemini-2.5-Pro & 99.9 & \textbf{91.7} & \textbf{95.4} & 21.5 & 69.2 & 86.4 & 62.2 & \textbf{23.0} \\
Gemini-2.5-Flash & 99.4 & 86.9 & 92.1 & 47.7 & 20.7 & 80.5 & \textbf{87.2} & 17.1 \\
\midrule
Claude-Haiku-4.5 & \textbf{100.0} & 89.7 & 87.4 & 97.6 & 68.1 & 71.3 & 51.7 & 3.6 \\
\midrule
Mistral-Large & 99.4 & 87.9 & 90.1 & 77.3 & 70.3 & 81.8 & 47.5 & 10.6 \\
Mistral-Small-24B & 99.1 & 82.0 & 85.9 & 97.8 & 49.3 & 0.6 & 44.8 & 2.6 \\
\midrule
Qwen3-235B & 99.9 & 88.3 & 92.6 & 67.4 & 71.2 & 70.9 & 50.0 & 2.8 \\
Qwen3-30B & 99.7 & 85.7 & 8.1 & \textbf{99.7} & 57.9 & 24.2 & 42.5 & 0.8 \\
Qwen3-14B & 99.5 & 83.2 & 5.3 & 98.2 & 52.4 & 18.6 & 38.2 & 2.2 \\
Qwen3-8B & 98.9 & 80.1 & 3.8 & 96.5 & 48.6 & 12.4 & 35.1 & 1.0 \\
\midrule
\textit{Mean} & \textit{99.5} & \textit{86.4} & \textit{75.4} & \textit{74.5} & \textit{62.8} & \textit{59.3} & \textit{51.8} & \textit{7.3} \\
\textit{Std} & \textit{0.8} & \textit{3.5} & \textit{32.1} & \textit{24.8} & \textit{15.8} & \textit{30.2} & \textit{13.6} & \textit{7.0} \\
\bottomrule
\end{tabular}
}
\end{table}

\begin{figure*}[htbp]
\centering
\includegraphics[width=\textwidth]{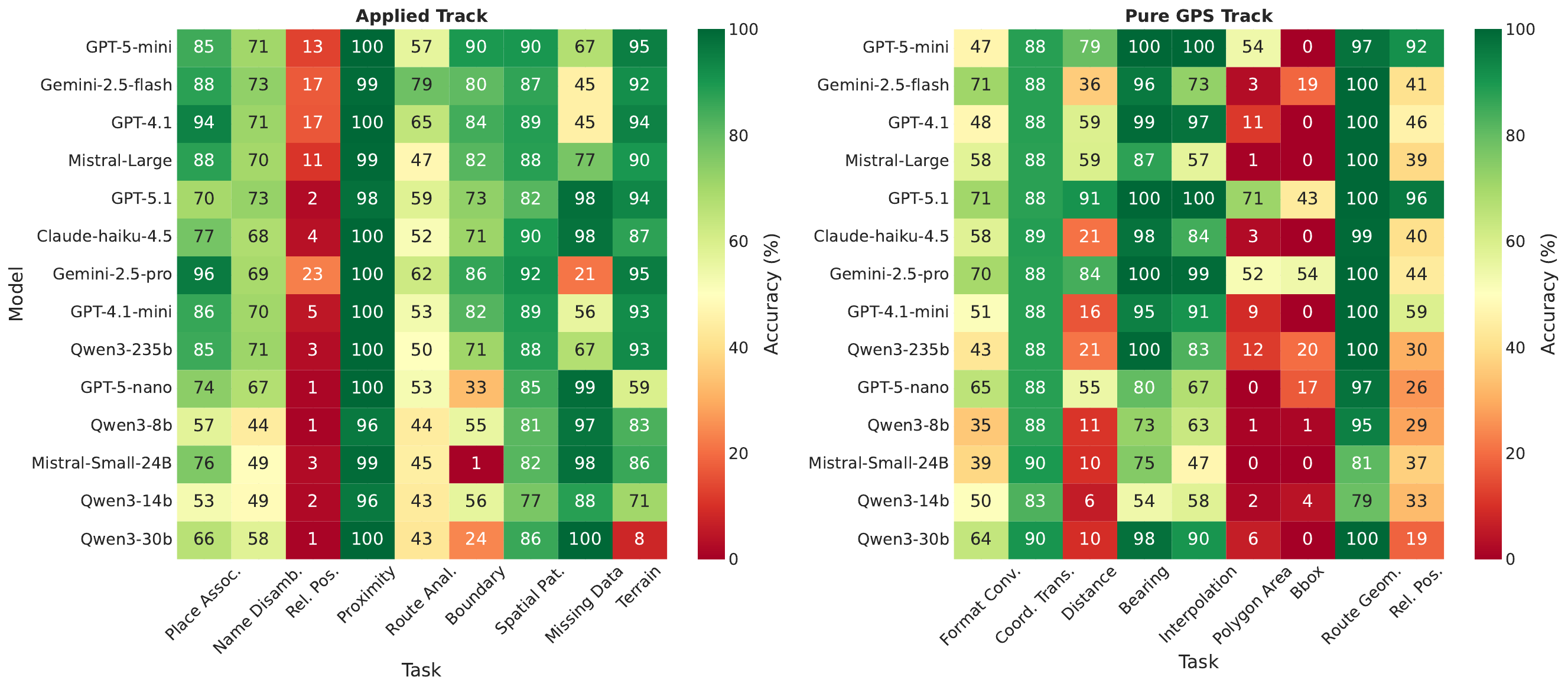}
\caption{Per-task accuracy (\%) across all models. (a) Applied Track tasks requiring world knowledge integration. (b) Pure GPS Track tasks requiring geometric computation. Green indicates high accuracy, red indicates low accuracy.}
\label{fig:heatmap_detailed}
\end{figure*}

\subsection{Per-Task Performance Breakdown}
\label{app:per_task}

Tables~\ref{tab:per_task_applied} and~\ref{tab:per_task_pure_gps} further reports per-task performance of the models over the two tracks, respectively, while Figure~\ref{fig:heatmap_detailed} visualizes these results. 

\begin{table}[htbp]
\centering
\caption{Per-task accuracy (\%) over Pure GPS Track tasks. Bold indicates best per task.}
\label{tab:per_task_pure_gps}
\setlength{\tabcolsep}{1pt}
\resizebox{\linewidth}{!}{
\begin{tabular}{lccccccccc}
\toprule
\textbf{Model} & \textbf{BBox} & \textbf{Format} & \textbf{Dist.} & \textbf{Bearing} & \textbf{Rel. Pos.} & \textbf{Route} & \textbf{Coord.} & \textbf{Interp.} & \textbf{Area} \\
\midrule
GPT-5.1 & 99.7 & 87.8 & \textbf{99.9} & \textbf{99.9} & \textbf{70.7} & \textbf{96.3} & \textbf{90.8} & \textbf{71.4} & 43.4 \\
GPT-5-mini & 97.5 & 88.0 & \textbf{99.9} & \textbf{99.8} & 46.6 & 91.8 & 79.4 & 54.1 & 0.0 \\
GPT-5-nano & 97.5 & 88.0 & 80.4 & 67.3 & 65.5 & 26.4 & 54.7 & 0.3 & 17.1 \\
GPT-4.1 & \textbf{100.0} & 88.1 & 99.1 & 97.5 & 47.5 & 46.0 & 59.0 & 11.4 & 0.0 \\
GPT-4.1-mini & 99.9 & 88.1 & 94.8 & 90.9 & 51.0 & 59.0 & 16.3 & 9.1 & 0.0 \\
\midrule
Gemini-2.5-Pro & \textbf{100.0} & 88.1 & \textbf{99.9} & 99.2 & 69.7 & 43.6 & 84.2 & 52.1 & \textbf{53.9} \\
Gemini-2.5-Flash & \textbf{100.0} & 88.0 & 96.3 & 72.7 & \textbf{70.8} & 41.0 & 36.1 & 2.7 & 18.9 \\
\midrule
Claude-Haiku-4.5 & 99.3 & 88.6 & 97.6 & 84.5 & 58.1 & 40.4 & 20.7 & 2.6 & 0.0 \\
\midrule
Mistral-Large & 99.7 & 87.8 & 87.2 & 56.7 & 57.6 & 39.0 & 59.5 & 0.7 & 0.0 \\
Mistral-Small-24B & 81.2 & 89.8 & 75.3 & 47.1 & 38.6 & 37.1 & 9.9 & 0.1 & 0.0 \\
\midrule
Qwen3-235B & 99.7 & 88.1 & 99.5 & 83.0 & 42.5 & 30.5 & 21.3 & 11.9 & 20.1 \\
Qwen3-30B & \textbf{100.0} & \textbf{90.4} & 97.8 & 90.5 & 64.1 & 18.7 & 9.7 & 6.4 & 0.0 \\
Qwen3-14B & 97.2 & 87.3 & 72.5 & 52.3 & 35.8 & 12.4 & 5.2 & 2.1 & 0.0 \\
Qwen3-8B & 95.8 & 86.1 & 68.2 & 45.1 & 32.4 & 8.6 & 3.8 & 0.8 & 0.0 \\
\midrule
\textit{Mean} & \textit{97.6} & \textit{88.3} & \textit{90.6} & \textit{77.6} & \textit{53.6} & \textit{42.2} & \textit{39.3} & \textit{16.1} & \textit{11.0} \\
\textit{Std} & \textit{5.0} & \textit{1.1} & \textit{12.1} & \textit{20.4} & \textit{13.8} & \textit{26.5} & \textit{31.2} & \textit{24.6} & \textit{18.1} \\
\bottomrule
\end{tabular}
}
\end{table}

\subsection{Qualitative Error Analysis}
\label{app:error_analysis}

We analyze model outputs across tasks to identify systematic error patterns. Table~\ref{tab:error_patterns} summarizes the dominant failure modes for a subset of tasks where clear patterns emerge.


\begin{table}[htbp]
\centering
\caption{Dominant error patterns by task category.}
\label{tab:error_patterns}
\small
\begin{tabular}{p{2.8cm}p{4.7cm}}
\toprule
\textbf{Task} & \textbf{Dominant Error Pattern} \\
\midrule
Place Association & Nearby city substitution \\
Polygon Area  & Formula stated but not executed \\
Coord. Interpolation & Cumulative numerical errors \\
Coord. Transformation & Nonlinear projection arithmetic \\
Bearing/Distance & Computational approximations \\
Spatial Patterns & Heuristic-based reasoning \\
Format Conversion & Precision mismatch \& hemisphere swaps \\ 
\bottomrule
\end{tabular}
\end{table}

\paragraph{Place association: nearby city substitution} 
Models consistently identify the correct country and province but substitute larger nearby cities for exact locations. For example, given coordinates 10.607°N, 72.979°W (Villanueva, Colombia), GPT-4.1 responds ``Maicao, La Guajira, Colombia'': the correct province but wrong city (Maicao is 50 km away). Similarly, coordinates for Emure-Ekiti, Nigeria yield ``Akure, Ondo State, Nigeria'' (the state capital). This pattern confirms that models encode coarse geographic knowledge but lack dense coordinate-to-city mappings:

\begin{quote}
\small
\textbf{Input:} What location is at coordinates 7.436403, 5.459255?\\
\textbf{Ground Truth:} Emure-Ekiti, Nigeria\\
\textbf{GPT-4.1:} ``The coordinates are in Nigeria. Specifically, this point is in the city of Akure, the capital of Ondo State.''\\
\textbf{Analysis:} Correct country and state, but substitutes state capital for actual city.
\end{quote}

\paragraph{Polygon area: formula without execution}
Models correctly identify the spherical excess formula but fail to execute the multi-step calculation precisely. Responses typically show abbreviated reasoning (``Approximate calculation yields ...'') followed by estimates that deviate significantly from ground truth:

\begin{quote}
\small
\textbf{Input:} Calculate the area of the polygon: Launceston, City of Port Phillip, Hervey Bay, Greenacre (Australia)\\
\textbf{Ground Truth:} 371,518 km\textsuperscript{2}\\
\textbf{GPT-4.1:} ``Use the spherical excess formula for a quadrilateral: Area $\approx$ E $\times$ R$^2$... Approximate calculation yields: FINAL ANSWER: 2,200,000 km\textsuperscript{2}''\\
\textbf{Analysis:} States correct method but estimate is 5.9$\times$ too large.
\end{quote}

\paragraph{Coordinate interpolation: cumulative numerical errors}
Models correctly apply the spherical linear interpolation (Slerp) algorithm conceptually but accumulate errors through multi-step calculations (coordinate conversion $\rightarrow$ Cartesian transformation $\rightarrow$ angle computation $\rightarrow$ weighting $\rightarrow$ interpolation $\rightarrow$ normalization $\rightarrow$ back-conversion). With tight tolerance (0.01°), even small intermediate errors compound:

\begin{quote}
\small
\textbf{Input:} Find the point 0.5 of the way from West Englewood (41.778°N, 87.667°W) to Achaguas (7.779°N, 68.224°W)\\
\textbf{Ground Truth:} 25.089°N, 76.560°W\\
\textbf{GPT-4.1:} 25.104°N, 76.484°W\\
\textbf{Analysis:} Latitude error: 0.015° ($>$ 0.01° tolerance), longitude error: 0.076°. Close but fails due to accumulated rounding in intermediate steps.
\end{quote}

\begin{table*}[htbp]
\centering
\caption{Accuracy (\%) by subregion across all GPSBench tasks. Bold indicates highest accuracy per task; `--' indicates insufficient data.}
\label{tab:regional_all_tasks_comprehensive}
\small
\begin{tabular}{lcccccccccc}
\toprule
\textbf{Task} & \textbf{N.Am} & \textbf{S.Am} & \textbf{W.Eur} & \textbf{E.Eur} & \textbf{E.Asia} & \textbf{S.Asia} & \textbf{SE.Asia} & \textbf{M.East} & \textbf{Africa} & \textbf{Oceania} \\
\midrule
\multicolumn{11}{l}{\textit{Applied Track}} \\
Place Assoc. & \textbf{14.7} & 6.7 & 11.0 & 8.3 & 3.8 & 2.6 & 5.9 & 5.0 & 4.5 & 8.0 \\
Name Disamb. & 99.1 & 99.5 & 99.1 & 98.8 & 98.8 & 97.1 & 99.3 & -- & 99.4 & \textbf{100.0} \\
Rel. Position & 48.1 & 55.4 & 53.7 & \textbf{59.7} & 44.6 & 56.5 & 53.1 & 53.4 & 55.8 & 56.5 \\
Proximity & 61.4 & 63.6 & 69.6 & \textbf{70.6} & 63.7 & 65.6 & 63.2 & 68.5 & 60.8 & 59.2 \\
Route Analysis & 85.8 & 81.8 & \textbf{91.4} & 88.6 & 80.9 & 86.2 & 84.3 & 88.8 & 88.2 & 84.0 \\
Boundary & 75.9 & 73.8 & \textbf{78.6} & 74.6 & 76.0 & 78.3 & 75.1 & 74.7 & 74.3 & 75.1 \\
Spatial Patterns & 81.4 & 84.4 & 76.9 & 80.4 & 82.6 & \textbf{87.1} & 82.9 & 76.1 & 78.0 & 83.1 \\
Terrain Class. & \textbf{69.8} & 69.3 & 50.3 & 50.0 & 52.4 & 58.8 & 62.6 & 55.0 & 60.4 & 61.6 \\
\midrule
\multicolumn{11}{l}{\textit{Pure GPS Track}} \\
Format Conv. & 99.5 & 97.4 & 99.3 & 99.5 & 98.9 & 99.1 & 98.9 & 99.5 & 97.8 & \textbf{100.0} \\
Coord. Trans. & \textbf{43.7} & 38.4 & 41.8 & 39.3 & 36.4 & 40.3 & 38.8 & 40.9 & 39.5 & 43.5 \\
Distance & 78.9 & 79.0 & 80.6 & 76.4 & 78.2 & 77.4 & 77.7 & \textbf{83.3} & 78.1 & 74.6 \\
Bearing & 68.6 & 75.8 & 69.2 & 67.4 & 62.3 & 72.7 & 72.4 & 73.7 & \textbf{79.4} & 64.4 \\
Interpolation & 14.2 & 17.9 & 16.7 & 14.9 & 11.2 & 15.4 & 11.6 & 17.1 & \textbf{21.7} & 8.6 \\
Polygon Area & 10.6 & \textbf{12.0} & 8.9 & 10.9 & 7.9 & 11.4 & 9.5 & 10.2 & 11.9 & 7.8 \\
Bounding Box & \textbf{98.5} & 96.7 & 98.0 & 97.8 & 97.9 & \textbf{98.5} & 96.4 & 96.1 & 95.9 & 95.2 \\
Route Geom. & 41.9 & 43.8 & 46.0 & 46.8 & 47.8 & 43.9 & \textbf{50.0} & 43.3 & 46.7 & 34.1 \\
Rel. Pos. (Pure) & 45.9 & 53.9 & 56.9 & 46.5 & 52.3 & \textbf{61.4} & 57.3 & 57.9 & 58.5 & 48.6 \\
\bottomrule
\end{tabular}
\end{table*}

\begin{figure*}[htbp]
\centering
\includegraphics[width=\textwidth]{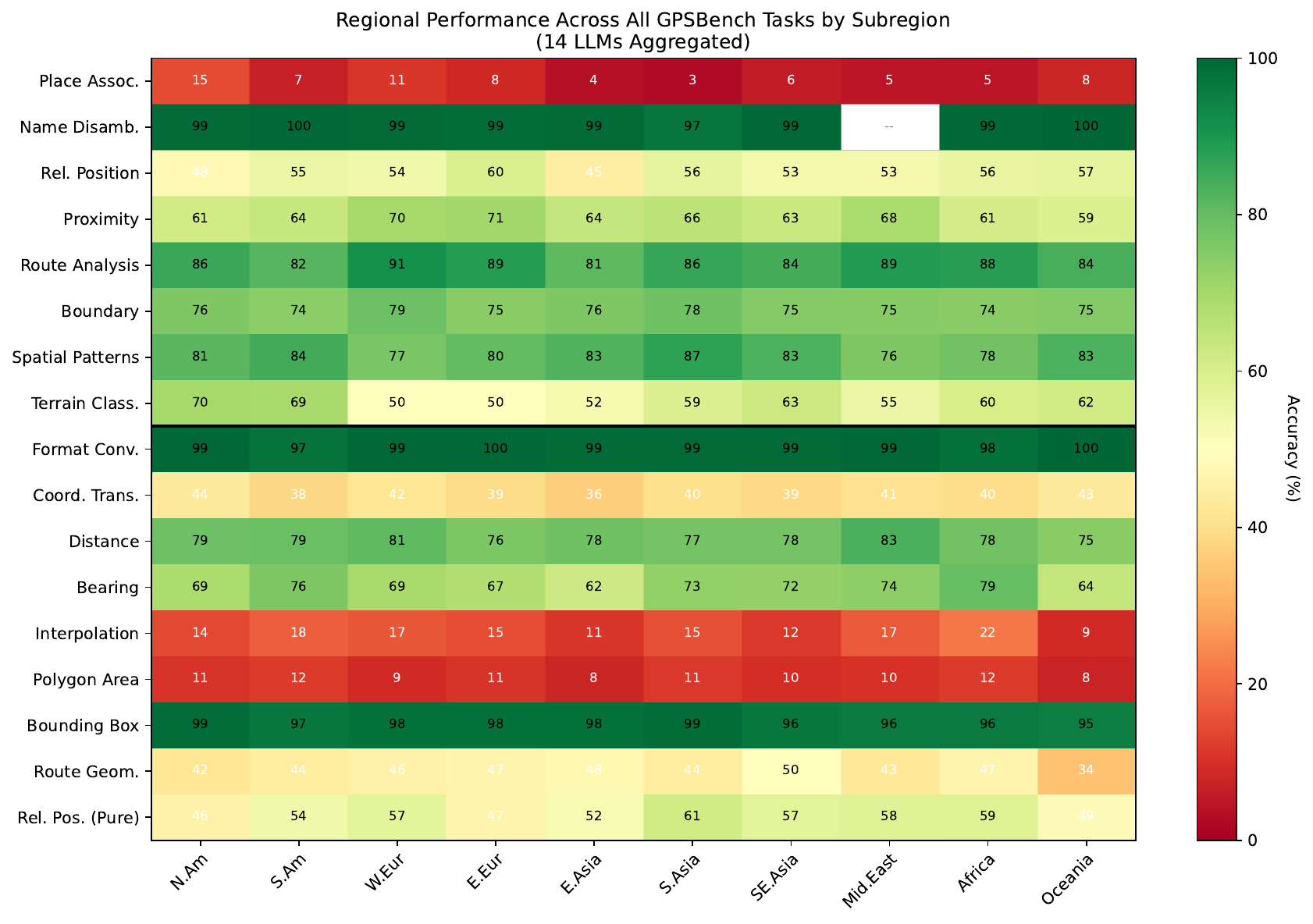}
\caption{Regional performance across all GPSBench tasks by subregion. Applied Track tasks (top) show varying degrees of geographic bias, with Place Association exhibiting the strongest disparity. Pure GPS Track tasks (bottom) show minimal regional variation, confirming that geometric computation is largely region-agnostic.}
\label{fig:regional_all_tasks_comprehensive}
\end{figure*}

\paragraph{Bearing computation: high success rate}
Bearing computation achieves high accuracy (up to 97.7\% for GPT-4.1) because it requires fewer computational steps than interpolation and has generous tolerance ($\pm$5°). The few failures occur on long-distance, cross-hemisphere routes where the great-circle path crosses the antimeridian:

\begin{quote}
\small
\textbf{Input:} Calculate bearing from Tacloban, Philippines (11.24°N, 125.00°E) to Cimarron Hills, USA (38.86°N, 104.70°W)\\
\textbf{Ground Truth:} 39.77° (NE)\\
\textbf{GPT-4.1:} [Explains formula but calculation incomplete]\\
\textbf{Analysis:} Cross-Pacific routes require careful handling of longitude wrapping; model's intermediate steps are correct but final computation not completed.
\end{quote}

\paragraph{Coordinate System Transformation: Nonlinear projection arithmetic}
Errors are due to projection complexity. UTM (Transverse Mercator) fails far more often than Web Mercator, and within Web Mercator, the linear $x$-formula ($x = R\lambda$) is usually correct while the nonlinear $y$-formula involving \texttt{tan} and \texttt{ln} accounts for most of the errors. This suggests models can handle linear arithmetic but struggle with chains of transcendental functions. Southern-hemisphere UTM is especially error-prone, as models sometimes truncate the 10-million-meter false northing offset:

\begin{quote}
\small
\textbf{Input:} Convert WGS84 coordinates for Ngong, Kenya (-1.3527°, 36.6699°) to UTM Zone 37S.\\
\textbf{Ground Truth:} Easting: 240,718 m, Northing: 9,850,361 m\\
\textbf{GPT-4.1:} Easting: 266,222 m, Northing: 985,045 m\\
\textbf{Analysis:} Easting is off by 25 km, and Northing drops a digit (985K instead of 9.85M), omitting the southern-hemisphere false northing.
\end{quote}

\paragraph{Spatial patterns: heuristic-based reasoning.}
When the geographic outlier is obvious (e.g., one city on a different continent), models succeed reliably. Errors occur when city names are ambiguous or when models apply simple geographic heuristics instead of computing centroid distances:

\begin{quote}
\small
\textbf{Input:} Which location is the geographic outlier: Hunts Cross, Atkarsk, Moskovskiy, Challans, Rye?\\
\textbf{Ground Truth:} Rye (Australia, 5,657 km from centroid)\\
\textbf{GPT-4.1:} ``Rye is a town in East Sussex, England... Atkarsk is the farthest east and most distant from the cluster of Western European locations. FINAL ANSWER: Atkarsk''\\
\textbf{Analysis:} Model incorrectly identifies Rye as the English town rather than Rye, Victoria, Australia. Uses ``farthest east'' heuristic instead of computing centroid distances, selecting Atkarsk (Russia, 3,068 km) over the true outlier.
\end{quote}

\paragraph{Format Conversion: precision mismatch and hemisphere swaps}
Format conversion achieves high accuracy (88\% for GPT-4.1), but the 121 errors reveal two distinct failure modes. The majority are precision mismatches: the model outputs correct values but with inconsistent decimal places (e.g., ``30.43'' instead of the expected ``30.4'') or omitted trailing zeros (e.g., ``30'' instead of ``30.0''). A smaller but more concerning class involves hemisphere label swaps, where latitude and longitude hemisphere indicators are transposed:

\begin{quote}
\small
\textbf{Input:} Convert 2.408500°, 42.977100° to DMS format.\\
\textbf{Ground Truth:} 2°24'30.6"N, 42°58'37.6"E\\
\textbf{GPT-4.1:} 2°24'30.6"E, 42°58'37.6"N\\
\textbf{Analysis:} Numerical conversion is exact, but N and E hemisphere labels are swapped between latitude and longitude, producing an invalid coordinate (2°E latitude, 42°N longitude).
\end{quote}

\subsection{Regional Performance Breakdown}
\label{app:regional_performance}

We analyze regional performance across tasks by extracting location information from coordinates, country codes, and prompt text. 
Table~\ref{tab:regional_all_tasks_comprehensive} reports the accuracy results across subregions, while Figure~\ref{fig:regional_all_tasks_comprehensive} visualizes the results. The results reveal that \textit{geographic bias is task-specific, not universal}.

\paragraph{Task-specific patterns} 
Place Association shows the strongest regional bias: North America (14.7\%) and Western Europe (11.0\%) outperform South Asia (2.6\%) and East Asia (3.8\%) by 4--6$\times$. This reflects sparse coordinate-to-city mappings in training data for non-Western regions. In contrast, most Pure GPS computation tasks show minimal regional variation: Distance (74.6--83.3\%), Bounding Box (95.2--98.5\%), and Format Conversion (97.4--100\%) are nearly uniform across subregions, confirming that geometric computation is region-agnostic. Name Disambiguation shows no bias (97.1--100\%), indicating that this task is ``solved'' regardless of location. The comprehensive analysis confirms that geographic bias stems primarily from knowledge gaps rather than computational limitations.

\subsection{Geographic Granularity Results}
\label{app:granularity_full}

Table~\ref{tab:place_assoc_granularity} reports Place association accuracy results over tests at different 
geographic granularity levels. 

\begin{table}[htbp]
\centering
\caption{Place Association accuracy (\%) by geographic granularity level across all 14 models.}
\label{tab:place_assoc_granularity}
\small
\begin{tabular}{lcccc}
\toprule
\textbf{Model} & \textbf{City} & \textbf{Province} & \textbf{Country} & \textbf{Region} \\
\midrule
GPT-5.1 & 12.9 & 61.2 & 81.1 & 0.4 \\
GPT-5-mini & 2.5 & 44.4 & 77.1 & 0.3 \\
GPT-5-nano & 1.3 & 32.2 & 71.9 & 0.2 \\
GPT-4.1 & 16.8 & 61.7 & 85.4 & 1.4 \\
GPT-4.1-mini & 4.7 & 49.1 & 82.1 & 2.4 \\
\midrule
Gemini-2.5-Pro & 23.0 & 72.9 & 96.5 & 40.0 \\
Gemini-2.5-Flash & 17.1 & 69.1 & 91.8 & 13.0 \\
\midrule
Claude-Haiku-4.5 & 3.6 & 48.9 & 86.0 & 38.4 \\
\midrule
Mistral-Large & 10.6 & 60.0 & 91.8 & 1.0 \\
Mistral-Small-24B & 2.6 & 42.9 & 86.1 & 7.0 \\
\midrule
Qwen3-235B & 2.8 & 52.5 & 86.5 & 2.8 \\
Qwen3-30B & 0.8 & 35.2 & 73.2 & 3.5 \\
Qwen3-14B & 2.2 & 29.8 & 63.4 & 0.1 \\
Qwen3-8B & 1.0 & 26.4 & 59.4 & 1.5 \\
\bottomrule
\end{tabular}
\end{table}

\subsection{Coordinate Noise Analysis}
\label{app:noise_analysis}

We test LLM robustness by adding Gaussian noise to coordinates at all granularity levels. This also probes memorization: if models memorized specific coordinate-to-place mappings from training data, perturbing coordinates would break these associations. Table~\ref{tab:noise_by_granularity} shows mean accuracy across models at each granularity level and noise magnitude. The relatively small performance variation at any level, including high-accuracy country identification (80\%), suggests models encode genuine geographic knowledge rather than memorized coordinate strings.

\begin{table}[htbp]
\centering
\caption{Mean accuracy (\%) by noise level and granularity. All levels show stable performance across noise magnitudes, suggesting generalized knowledge rather than memorization.}
\label{tab:noise_by_granularity}
\small
\begin{tabular}{lccc}
\toprule
\textbf{Noise} & \textbf{Country} & \textbf{Province} & \textbf{City} \\
\midrule
Clean & 80.7 & 46.3 & 7.4 \\
10m & 79.1 ($-$1.6) & 51.8 (+5.5) & 5.8 ($-$1.6) \\
50m & 80.1 ($-$0.5) & 52.1 (+5.8) & 9.3 (+1.9) \\
100m & 82.3 (+1.6) & 45.8 ($-$0.5) & 7.1 ($-$0.3) \\
500m & 82.1 (+1.5) & 49.7 (+3.4) & 7.0 ($-$0.5) \\
1km & 81.0 (+0.4) & 48.3 (+2.0) & 7.0 ($-$0.4) \\
\midrule
\textit{Max $|\Delta|$} & \textit{1.6} & \textit{5.8} & \textit{1.9} \\
\bottomrule
\end{tabular}
\end{table}

Table~\ref{tab:place_assoc_noise_results} details per-model city-level accuracy changes.

\begin{table}[htbp]
\centering
\caption{Place association accuracy change ($\Delta$\%) relative to clean coordinates across all 14 models. Values show difference from clean baseline. Note: no correlation between noise magnitude and $|\Delta|$ (mean $|\Delta|$ = 1.3--2.1\% across all levels).}
\label{tab:place_assoc_noise_results}
\setlength{\tabcolsep}{1pt}
\resizebox{\linewidth}{!}{
\begin{tabular}{lccccc|c}
\toprule
\textbf{Model} & \textbf{Clean (\%)} & \textbf{$\Delta$10m} & \textbf{$\Delta$50m} & \textbf{$\Delta$100m} & \textbf{$\Delta$500m} & \textbf{$\Delta$1km} \\
\midrule
GPT-5.1 & 13.1 & $-$2.2 & +3.7 & $-$2.0 & $-$0.4 & $-$0.5 \\
GPT-5-mini & 1.7 & $-$1.1 & +0.1 & +2.2 & +2.5 & +0.8 \\
GPT-5-nano & 1.1 & $-$1.1 & +1.9 & +0.2 & $-$1.1 & +0.9 \\
GPT-4.1 & 16.0 & $-$2.7 & +5.7 & $-$1.5 & +0.4 & +2.3 \\
GPT-4.1-mini & 5.7 & $-$3.9 & +0.9 & +0.9 & $-$0.9 & $-$2.7 \\
\midrule
Gemini-2.5-Pro & 24.0 & $-$1.0 & +4.9 & $-$3.6 & $-$3.4 & $-$2.7 \\
Gemini-2.5-Flash & 18.9 & $-$0.7 & $-$0.2 & $-$5.0 & $-$0.1 & $-$4.6 \\
\midrule
Claude-Haiku-4.5 & 4.6 & $-$4.0 & +1.5 & $-$0.6 & $-$0.9 & $-$1.5 \\
\midrule
Mistral-Large & 13.1 & $-$4.7 & $-$1.7 & $-$2.0 & $-$4.7 & $-$2.5 \\
Mistral-Small-24B & 1.7 & $-$1.1 & +1.9 & +2.2 & +0.7 & +1.8 \\
\midrule
Qwen3-235B & 2.9 & $-$1.0 & +0.8 & $-$0.2 & +0.8 & $-$0.3 \\
Qwen3-30B & 0.6 & $-$0.6 & +1.2 & +0.7 & $-$0.6 & +0.4 \\
Qwen3-14B & 0.6 & +0.6 & +3.0 & +2.1 & +1.2 & +2.5 \\
Qwen3-8B & 0.0 & +0.6 & +2.4 & +2.6 & +0.0 & +0.5 \\
\midrule
\textit{Mean $|\Delta|$} & -- & \textit{1.8} & \textit{2.1} & \textit{1.8} & \textit{1.3} & \textit{1.7} \\
\bottomrule
\end{tabular}
}
\end{table}

\subsection{GPS Augmentation Detailed Results}
\label{app:downstream}

Table~\ref{tab:gps_augmentation_downstream} reports accuracy results when GPS coordinates are added to downstream tasks. 

\begin{table}[htbp]
\centering
\caption{Effect of GPS augmentation on downstream tasks (\%). Results using GPT-4.1.}
\label{tab:gps_augmentation_downstream}
\setlength{\tabcolsep}{1pt}
\small
\begin{tabular}{p{2cm}p{2.5cm}cccc}
\toprule
\textbf{Dataset} & \textbf{Category} & \textbf{N} & \textbf{Base} & \textbf{+GPS} & \textbf{$\Delta$} \\
\midrule
\multirow{4}{*}{\shortstack[l]{MapEval}}
    & Trip planning & 11 & 90.9 & 100.0 & +9.1 \\
    & POI queries & 51 & 72.5 & 76.5 & +3.9 \\
    & Nearby search & 3 & 66.7 & 100.0 & +33.3 \\
    \cmidrule{2-6}
    & \textit{Overall} & \textit{66} & \textit{75.8} & \textit{81.8} & \textit{+6.1} \\
\midrule
\multirow{5}{*}{\shortstack[l]{Hierarchical\\Spatial}}
    & Hierarchical bias & 10 & 70.0 & 100.0 & +30.0 \\
    & Alignment bias & 4 & 50.0 & 100.0 & +50.0 \\
    & Proximity bias & 4 & 0.0 & 0.0 & 0.0 \\
    & Rotation bias & 4 & 0.0 & 0.0 & 0.0 \\
    \cmidrule{2-6}
    & \textit{Overall} & \textit{22} & \textit{40.9} & \textit{63.6} & \textit{+22.7} \\
\bottomrule
\end{tabular}
\end{table}

\subsection{Finetuning Analysis}
\label{app:finetuning}

\paragraph{Finetuning setting}
We finetune Qwen3-30B-A3B-Instruct (a 30B-total, 3B-active MoE model) using LoRA \cite{hulora2022} on the GPSBench training split. Training uses both the Pure GPS and Applied tracks. We use the Tinker platform \citep{tml2025tinker} for distributed finetuning with the following hyperparameters: learning rate $2 \times 10^{-5}$ with linear decay, batch size 32, LoRA rank 64, maximum sequence length 16{,}384 tokens, and 2 training epochs. The low learning rate is chosen to mitigate catastrophic forgetting of pretrained knowledge. Training data consists of the standard 60\% train split (2{,}040 samples per task $\times$ 17 tasks = 34{,}680 total samples). Evaluation is performed on the 30\% held-out test split. 

\begin{table}[htbp]
\centering
\caption{Overall finetuning results on Qwen3-30B.}
\label{tab:finetuning_overall}
\small
\begin{tabular}{lccc}
\toprule
\textbf{Track} & \textbf{Zero-shot} & \textbf{Finetuned} & \textbf{$\Delta$} \\
\midrule
Applied & 52.3\% & 50.7\% & $-$1.6\% \\
Pure GPS & 53.1\% & 57.4\% & +4.3\% \\
\midrule
\textit{Combined} & \textit{52.7\%} & \textit{54.1\%} & \textit{+1.5\%} \\
\bottomrule
\end{tabular}
\end{table}

Table~\ref{tab:finetuning_per_task} provides per-task breakdown, revealing divergent effects on geometric vs.\ world-knowledge tasks.

\begin{table}[htbp]
\centering
\caption{Per-task finetuning comparison on Qwen3-30B (\%). Green: improvement ($>$5\%), red: degradation ($<$$-$5\%).}
\label{tab:finetuning_per_task}
\small
\begin{tabular}{lccc}
\toprule
\textbf{Task} & \textbf{Zero-shot} & \textbf{Finetuned} & \textbf{$\Delta$} \\
\midrule
\multicolumn{4}{c}{\textit{Applied Track}} \\
\midrule
Place Association & 0.8 & 1.7 & +0.9 \\
Name Disambiguation & 99.7 & 82.6 & \textcolor{red}{$-$17.1} \\
Relative Position & 42.5 & 22.9 & \textcolor{red}{$-$19.6} \\
Proximity & 24.2 & 15.4 & \textcolor{red}{$-$8.8} \\
Route Analysis & 85.7 & 89.9 & +4.2 \\
Boundary Analysis & 99.7 & 74.5 & \textcolor{red}{$-$25.2} \\
Spatial Patterns & 8.1 & 64.6 & \textcolor{green!60!black}{+56.5} \\
Terrain & 57.9 & 53.9 & $-$4.0 \\
\midrule
\textit{Applied Overall} & \textit{52.3} & \textit{50.7} & \textit{$-$1.6} \\
\midrule
\multicolumn{4}{c}{\textit{Pure GPS Track}} \\
\midrule
Format Conversion & 90.4 & 90.9 & +0.5 \\
Coord.\ Transform & 9.7 & 6.6 & $-$3.1 \\
Distance & 97.8 & 94.2 & $-$3.6 \\
Bearing & 90.5 & 78.2 & \textcolor{red}{$-$12.3} \\
Interpolation & 6.4 & 25.3 & \textcolor{green!60!black}{+18.9} \\
Polygon Area & 0.0 & 25.9 & \textcolor{green!60!black}{+25.9} \\
Bounding Box & 100.0 & 99.9 & $-$0.1 \\
Route Geometry & 18.7 & 31.1 & \textcolor{green!60!black}{+12.4} \\
Relative Position & 64.1 & 64.1 & 0.0 \\
\midrule
\textit{Pure GPS Overall} & \textit{53.1} & \textit{57.4} & \textit{+4.3} \\
\bottomrule
\end{tabular}
\end{table}



\end{document}